\documentclass[twocolumn]{svjour3}          
\smartqed  
\usepackage{amsmath}
\usepackage{color}
\usepackage{graphicx}
\usepackage{subfigure}
\usepackage{multirow}
\usepackage{caption}
\usepackage{threeparttable}
\usepackage{algorithm}
\usepackage{algorithmic}
\usepackage{makecell}
\usepackage{colortbl}
\usepackage[table]{xcolor}
\usepackage[nocompress]{cite}
\newcommand{\ve}[1]{\mbox{{\bf #1}}}
\newcolumntype{L}[1]{>{\raggedright\let\newline\\\arraybackslash\hspace{0pt}}m{#1}}
\newcolumntype{C}[1]{>{\centering\let\newline\\\arraybackslash\hspace{0pt}}m{#1}}
\newcolumntype{R}[1]{>{\raggedleft\let\newline\\\arraybackslash\hspace{0pt}}m{#1}}

\begin{document}

\title{Color Constancy with Derivative Colors
}


\author{Huan Lei         \and
        Guang Jiang  \and 
         Long Quan
}


\institute{Huan Lei \at
State Key Laboratory of Integrated Services Networks, School of Telecommunication
Engineering, Xidian University, Xi'an, 710071, P.R. China. \\
              \email{hlei.ziyan@gmail.com}\\
           \and
           Guang Jiang \at
State Key Laboratory of Integrated Services Networks, School of Telecommunication
Engineering, Xidian University, Xi'an, 710071, P.R. China. \\
\email{gjiang@mail.xidian.edu.cn}
\and
Long Quan\at
The Department of Computer Science and Engineering, Hong Kong University of
Science and Technology, Clear Water Bay, Kowloon, Hong Kong, China. \\
\email{quan@cse.ust.hk}
}

\date{Received: date / Accepted: date}

\maketitle

\begin{abstract}
Information about the illuminant color is well contained in both achromatic regions and the
specular components of highlight regions.
 In this paper, we propose a novel way to achieve color constancy by exploiting such clues.
The key to our approach lies
in the use of suitably extracted derivative colors, which are able to compute the illuminant color robustly with kernel density estimation.
While extracting derivative colors from achromatic regions to approximate the illuminant color well is basically straightforward, the success of our extraction in highlight regions
is attributed to
the different rates of variation of the diffuse and specular magnitudes in the dichromatic reflection model.
The proposed approach requires no training phase and is simple to implement. More significantly, it performs quite satisfactorily under inter-database parameter settings.
Our experiments on three standard databases demonstrate its effectiveness and fine performance in comparison to state-of-the-art methods.
\end{abstract}

\section{Introduction}
\label{intro}
Human visual system inherently has the ability of color constancy. However, for images
captured by digital cameras, the object colors are shifted by variations of the illuminant.
In practice, many high-level computer vision applications demand color constancy as a
preprocessing step to ensure that the same object color
taken under different illuminants can be accurately matched to their canonical ones.
The most significant part of computational color constancy is to estimate
the illuminant color.

Many color constancy algorithms have been presented
so far. According to whether the parameter setting of an algorithm is kept fixed or not, they can be classified into
the static and the learning-based groups \cite{gijsenij2011computational}. While, according to whether an algorithm
depends on understanding the physical process of the reflected light, color constancy approaches can alternatively be categorized into the physics-based and the statistics-based groups mainly \cite{finlayson2001solving}, which is the very categorization we adopt in this paper.
Although the statistics-based methods are accused of error-prone with limited object colors (e.g., \cite{forsyth1990novel, finlayson2001color,cardei2002estimating}) while
the physics-based ones are not in theory, existing physics-based methods are nearly altogether inferior to the statistics-based ones, particularly due to their mediocre performance on the standard databases. Nonetheless, the physics-based methods have such advantages such as fast execution, requiring few parameters and no training phase \cite{gijsenij2011computational}, which continually attract researchers in this field to exploit physics-based information for better color constancy algorithms.

Widely among the physics-based methods, the dichromatic reflection
model \cite{shafer1985using} has been explored to achieve color constancy.
Two dominant reasons contribute to this choice.
Firstly, the dichromatic reflection model provides a fine description about the imaging process
compared with the traditional Lambertian model. Secondly,
under the neutral interface reflection (NIR) assumption \cite{lee1986method}, the specular components of highlight pixels will perfectly contain information about the illuminant color.
The earlier methods solve for color constancy with intersection of dichromatic lines formed by different object
colors (e.g., \cite{lee1986method,tominaga1996multichannel,finlayson2001convex}).
Yet, they are rarely functional outside the lab since a challenging pre-segmentation of different object colors is demanded in the highlight regions.
In addition, they are inapplicable to scenes with uniform object color because of the
absence of intersection. Then, the
Planckian locus is introduced as a constraint which makes estimation
from uniform object color possible \cite{finlayson2001solving}.
Later, the inverse-intensity chromaticity (IIC) space is defined to estimate the illuminant color
\cite{tan2004color}, which eliminates the requirement for pre-segmentation. Although the approach is suitable for both uniform and highly textured surfaces, its performance is unacceptable on the standard databases. There are also some methods estimating the illuminant color from intersection of
dichromatic planes formed by different object colors (e.g., \cite{toro2007multilinear, shi2008dichromatic, toro2008dichromatic}), which are therefore inapplicable to uniform object colors as well. In addition, the identification of different
dichromatic planes is rather difficult.
Recently, geometric mean of highlight pixels is suggested to be taken as the illuminant color \cite{drew2014zeta}, which
again avoids pre-segmentation. More excitingly, the approach improves the estimation accuracy reported by the physics-based methods in a large degree.

Similar to the specular components of highlight pixels,
pixels whose albedos are achromatic also contain perfect reflectance of the illuminant, whether they are diffuse or with specular components.
In this paper, we exploit to achieve color constancy for a single image based
on both achromatic regions and specularity in highlight regions.
Although estimating the illuminant color in a pixel level is effective in achromatic regions, it is difficult to extract information about the illuminant color
from highlight regions. As a result, we propose to estimate the illuminant color using derivative colors, which are able to successfully combine the illuminant information from both achromatic and highlight regions.
While
extracting derivative colors from achromatic regions to approximate the illuminant color well is basically straightforward, the success of
our extraction in highlight regions is attributed to different rates of variation of the diffuse and specular magnitudes in the dichromatic
reflection model. Thus, compare to the physics-based methods presented purely according to the Lambertian model (e.g., \cite{geusebroek2001color, geusebroek2002physical}), the proposed approach exploits the significant specular information as well.
On the other hand,
superior to the dichromatic-line-based or dichromatic-plane-based methods,
the proposed approach readily overcomes the challenge of pre-segmenting different object colors
and is applicable to both uniform and highly textured scenes.
Meanwhile, better than IIC, which demands enough highlight pixels of the same object color to be with constant diffuse magnitude,
the proposed approach eliminates this inappropriate constraint and dependence on the diffuse component.
More importantly, extracting the derivative colors with suitably spatial operators, the proposed approach is able to estimate the illuminant color robustly with kernel density estimation. Similar to the previous physics-based methods, the proposed approach requires no any training session. Yet, it overcomes their
drawbacks such as difficult to implement and mediocre performance.

 Indeed, we are not the first to perform the estimation from derivative structures of an image. Methods exploring derivative structures of an image exist in the statistics-based group (e.g., \cite{van2007edge, gijsenij2010generalized, chakrabarti2012color, finlayson2013corrected}).
 van de Weijer et al. proposed the gray-edge method which assumes the
 average reflectance in the derivative structure of an image to be achromatic \cite{van2007edge}. One drawback of the approach is that it uses
 information from derivative structure of an image indiscriminately. After all, not derivatives of all pixels are helpful for estimating
 the illuminant color.
 Another drawback of the approach should be its inferior inter-database performance. That is, when the optimal parameter setting from one
 database is applied to another, its performance becomes mediocre.
Then, two interesting works based on the gray-edge method suggest that the derivative information should be utilized selectively \cite{gijsenij2012improving, joze2012role}. In particular,
Gijsenij et al. experimentally showed that
performance of the gray-edge method can be improved significantly if contributions of specular edges are weighted higher \cite{gijsenij2012improving}.
One thing which should be noted explicitly is that their edge classification mechanism according to \cite{van2005edge} makes
edges of achromatic pixels to be both specular and shadow.
vaezi Joze et al. extended the white-patch assumption \cite{land1977retinex} to bright pixels and alternatively
showed that constraining the gray-edge method to the bright pixels can obtain better color constancy results \cite{joze2012role}. Nevertheless, identical
to the original gray-edge method, they still suffer from the problem of unsatisfactory performance under inter-database configurations.
In contrast, to the best of our knowledge,
  we are the first to analyze the physical feasibility of estimating the illuminant color from derivative structure of an image in the chromaticity space, which shows more
 in-depth explanation for the improved performance achieved by \cite{gijsenij2012improving, joze2012role}.
Superior to \cite{van2007edge, gijsenij2012improving, joze2012role}, the proposed approach solves the problem
in the chromaticity space with the nonparametric kernel density estimation, eliminating the requirements for carefully parameter tuning.
As a matter of fact, it performs rather satisfactorily under inter-database parameter settings.
As to the rest statistics-based methods which solves color constancy from the
derivative structure of an image, they altogether require training data to be available (i.e., \cite{gijsenij2010generalized, chakrabarti2012color, finlayson2013corrected}). The proposed approach, despite requiring no training session, performs either competitive or better than them.
 Cheng et al. attributed the effectiveness of these derivative-based methods to color difference
 according to purely experimental feedback \cite{cheng2014illuminant}.
Yet, based on our analysis, such success is due to the extraction of derivative colors
  from the achromatic and highlight regions.
We conduct experiments on three linear standard databases: the SFU laboratory database \cite{barnard2002comparison}, the Gehler-Shi database \cite{shi2010data}, and the SFU HDR database \cite{funt2010rehabilitation}.
 Our experimental results demonstrate the effectiveness and exciting performance of the proposed approach in comparison to state-of-the-art methods.
\section{Related Work}
We review firstly methods in the physics-based group, focusing on methods based on the dichromatic reflection model,
and then methods in the statistics-based group.

Klinker et al. suggested to extract a T-shape color distribution from some uniform object color in the RGB space \cite{klinker1988measurement}, which is rather difficult and unreliable when applied to real images. Lee proposed to estimate the illuminant color by detecting intersections of multiple dichromatic lines in the chromaticity space \cite{lee1986method}. Then, many approaches were introduced to improve its performance \cite{tominaga2002natural,tominaga1989standard,tominaga1996multichannel,lehmann2001color,finlayson2001convex}. In particular, Finlayson and Schaefer imposed
a constraint on the illuminant color based on statistics of natural illuminate colors \cite{finlayson2001convex}.
However, since these approaches rely on intersections of different dichromatic lines, they are altogether infeasible for images with uniform object color.
Nevertheless, the chromaticities of common light sources
altogether locate closely to the Planckian locus of black-body radiators as shown in \cite{finlayson2001solving, mazin2012illuminant}. Therefore, Finlayson
and Schaefer further utilized the Planckian locus as a constraint to address the above mentioned issue and solved the
problem in theory \cite{finlayson2001solving}.
%
However, applying this method to images with multiple object colors demands a challenging pre-segmentation
to acquire clusterings of uniform object color.
Tan et al.
introduced an inverse-intensity chromaticity (IIC) space to estimate the illuminant color
using hough transformation \cite{tan2004color}. Although the approach is
suitable for both uniform and highly textured surfaces, its performance is unacceptable
on the standard databases. Toro and
Funt presented a multi-linear
constraint on the illuminant color with several dichromatic
planes \cite{toro2007multilinear}, which requires the representative colors of any given material to be identifiable.
And the identification is fulfilled with generalized principal component analysis in \cite{toro2007multilinear} and with mean shift in \cite{toro2008dichromatic}.
Shi et al. relaxed such requirements with a voting procedure that involves performing the hough transform twice \cite{shi2008dichromatic}.
However, this voting procedure still estimates color of the scene illuminant based on the intersection of different
dichromatic planes, so it will cease to be effective for uniform surfaces.
Meanwhile, using the hough transform twice will inevitably make the estimation rather time-consuming.
All of these approaches, despite with reasonable theoretical derivations,
either are difficult to implement or show mediocre performance on real images, which
significantly limit their applications in practice.
Recently, Drew et al.
proposed to estimate the illuminant color from a novel feature
named the zeta-image \cite{drew2014zeta}. They detected specular pixels with a planar constraint and improved the estimation accuracy
in a large degree. In contrast,
Prinet et al. proposed to extract specular information from the temporal derivative structure of
an image in video sequences and estimate the illuminant color with MAP \cite{prinet2013illuminant}.
Nevertheless, it is inapplicable for color constancy from a single image.

As to the statistics-based group, they can be further categorized into two dominant subgroups:
the derivative-based methods and the pixel-based methods.
Since the derivative-based methods
are more consistent with our work, we analyze them in the first place, and then
the pixel-based methods.

van de Weijer et al. generalized the earlier gray assumptions to
derivative structures of an image and presented the gray-edge method \cite{van2007edge},
whose performance depends highly on the carefully tuned parameters.
Generalized gamut mapping \cite{gijsenij2010generalized}, which builds the gamut from the
derivative structure of the image rather than colors are originally
present in the image, performs more stably than the original algorithm \cite{forsyth1990novel}.
Chakrabarti et al. explored to solve color constancy
by modeling the spatial-spectral statistics of an image originally with the gaussian function \cite{chakrabarti2008color}.
Later, they improved the performance by replacing the gaussian function with the heavy-tailed radial exponential function \cite{chakrabarti2012color}, and report satisfactory results on the Gehler-Shi database \cite{shi2010data}. Yet, the performance of their approach depends highly on the training subsets selected from the database.
Finlayson introduced the corrected-moment approach
by adding a correction step to the gray-based methods \cite{finlayson2013corrected}.
Although the approach reports better experimental
results on the standard databases, it requires the correct illuminant colors to be provided as candidates.
While these methods generally use information from the derivative structure of an image indiscriminately, two interesting works based on the gray-edge method \cite{van2007edge} show that the derivative information extracted from different kinds of pixels can influence the estimation results differently \cite{gijsenij2012improving, joze2012role}. In particular, Gijsenij et al. experimentally showed that
performance of the original gray-edge method \cite{van2007edge} could be improved significantly when the estimation is accomplished based on specular edges in \cite{gijsenij2012improving}. And they detected the specular edges using an iterative weighting mechanism. However, the optimal parameter settings of this approach vary seriously among different databases. vaezi Joze et al. extended the white-patch assumption \cite{land1977retinex} to bright pixels and alternatively
showed that constraining the original gray-edge method to bright pixels can improve its performance significantly \cite{joze2012role}.
Yet, their approach still suffers from the problem of inferior inter-database performance, similar to the methods \cite{gijsenij2012improving, van2007edge}.

Besides, many algorithms based on pixel level information exist in the literature of color constancy.
The low-level statistics-based methods estimate the illuminant color based on
simple statistical assumptions. For example, the gray-world hypothesis
assumes the average reflectance in
a scene under a neutral light source to be achromatic \cite{buchsbaum1980spatial}.
The white-patch hypothesis assumes that the maximum response
in the RGB-channels is caused by a perfect reflectance \cite{land1977retinex}. Shades of gray
estimates the illuminant color from a more general Minkowski framework \cite{finlayson2004shades}.
Despite the simplicity of these low-level statistics-based methods,
their performance is far from satisfactory even under optimal parameter settings.
The gamut mapping
algorithm represents the object colors that can be observed under a
canonical illuminant with the canonical gamut, and by mapping gamut of the input image
to the canonical gamut, the illuminant color can be therefore estimated \cite{forsyth1990novel}. Yet,
its performance tends to be error-prone when the number of different object colors is limited.
In addition, the canonical gamut depends on the training subsets selected from the database as well.
Color by correlation \cite{finlayson2001color} is
actually a discrete implementation of gamut mapping, but it can formulate many other algorithms
into its framework.
Alternatively, the bayesian approaches
\cite{gehler2008bayesian,brainard1997bayesian,rosenberg2003bayesian} model the
variability of surface reflectance and the illuminant color as random variables, and then estimate
the illuminant color from the posterior distribution conditioned on image intensity data.
In addition, introducing a representation of binarized chromaticity
histograms, the illuminant color can be recovered with either
neural networks \cite{cardei2002estimating} or support vector regression \cite{funt2004estimating}.
However, the performance of these approaches on the standard databases are not outstanding.
 Cheng et al. attributed the effectiveness of derivative-based methods to color difference
 according to purely experimental feedback \cite{cheng2014illuminant},
 and proposed to estimate the illuminant color through a principle component analysis on bright and dark pixels
\cite{cheng2014illuminant}. The performance of their approach is generally good on standard databases.
Chakrabarti recently proposed to estimate the illuminant color by training a luminance-to-chromaticity
classifier \cite{chakrabarti2015color}.

There are also some methods exploring complex information to achieve color constancy
among the statistics-based group (e.g., \cite{gijsenij2011color, bianco2008improving, van2007using}).
Most recently,
vaezi Joze and Drew presented an exemplar-based approach,
which estimated the illuminant color by using the color and texture feature cues to perform a nearest neighbour search
among the training data \cite{vaezi2014exemplar}.
They provided state-of-the-art results on standard databases. Yet, when this approach was applied
to inter-database settings, its performance degraded dramatically.
Cheng et al. presented to estimate the illuminant color quite fast using
an ensemble of regression trees, which is trained with four kinds of simple color feature cues
\cite{cheng2015effective}.
More generally, Li et al. proposed such an approach, which estimates the illuminant color by training
a sparse representation for multiple feature cues obtained from
different color constancy approaches and achieves fairly good results \cite{li2015multi}. Yet, the
approach is rather complicated and meanwhile time-consuming to be executed.

Methods beyond the physics-based and statistics-based groups are limited in number. Gao et al.
presented to fulfill color constancy using double-opponency
from a biological standpoint \cite{gao2015color}.
This method performs comparable to the complex ones under optimal parameter settings on different databases. Meanwhile, it shows
 better inter-database performance than the exemplar-based approach \cite{vaezi2014exemplar}.
\section{Approach}
\label{Approach}
Since achromatic regions can be utilized readily to estimate the illuminant color, our analysis is mainly focused on color constancy from the specular components in highlight regions. We introduce the well-known dichromatic reflection model in Section \ref{Basics}, validate the theoretical feasibility of using derivative colors extracted from highlight regions to estimate the illuminant color in Section \ref{theory}.
Yet, despite its simplicity, we also concisely analyze derivative colors from achromatic regions in the last part of Section \ref{theory}.
Finally in Section \ref{practice}, we design an algorithm which is quite robust for real images.
\subsection{The Dichromatic Reflection Model}
\label{Basics}
For a scene illuminated by a single illuminant, its spectral distribution is usually considered uniform.
Highlights of inhomogeneous dielectric objects are linear combinations of the diffuse and specular components
according to the dichromatic reflection model \cite{shafer1985using}. Based on the neutral interface
reflection (NIR) assumption \cite{lee1990modeling}, the interface reflection spectrum will be the same as that
of the illuminant. We formulate the dichromatic model depending on this assumption.
A pixel in an image of inhomogeneous
dielectric object taken by a digital color camera can be expressed as
\begin{equation}\label{equ1}
\ve{I}(\ve{x}) =  w_d(\ve{x})\int_\Omega e(\lambda)g(\lambda,\ve{x})\ve{q}(\lambda)d\lambda
+w_s(\ve{x})\int_\Omega e(\lambda)\ve{q}(\lambda)d\lambda,
\end{equation}
in which $e(\lambda)$ is the relative illumination spectrum, $g(\lambda,\ve{x})$ is
the spectral reflectance of the material, and $\ve{q}(\lambda)$ is a vector with its elements representing
the sensitivity functions of the
camera's corresponding channels. In addition, $\Omega$ refers to the visible spectrum. $w_d(\ve{x})$, $w_s(\ve{x})$
are respectively the magnitudes of body reflection and interface reflection.
More concisely, Eq. (\ref{equ1})
can be simplified as
\begin{equation}\label{BRDF}
\ve{I}(\ve{x})= m_d(\ve{x})\ve{D}(\ve{x})+m_s(\ve{x})\ve{S},
\end{equation}
in which $m_d(\ve{x})$ and $m_s(\ve{x})$ denote magnitudes of the diffuse reflection
and specular reflection ordinally.
$\ve{D}(\ve{x})=[D_r(\ve{x})\ D_g(\ve{x})\ D_b(\ve{x})]^T$ and $\ve{S}=[S_r\ S_g\ S_b]^T$ correspond respectively to
the intrinsic object color and the illuminant color.
Without loss of any generality, we constrain
\begin{equation}
    \sum_k D_k(\ve{x})=1,    \sum_k S_k=1, k\in\{r,g,b\}.
\end{equation}

\begin{figure}[!t]
  \centering
  \subfigure[] {\label{synimg}\includegraphics[width=24mm]{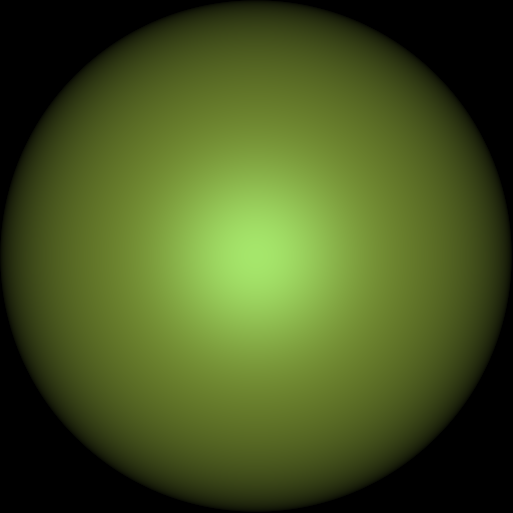}}  \hspace{-1.6mm}
  \subfigure[] {\label{synimg_md}\includegraphics[width=24mm]{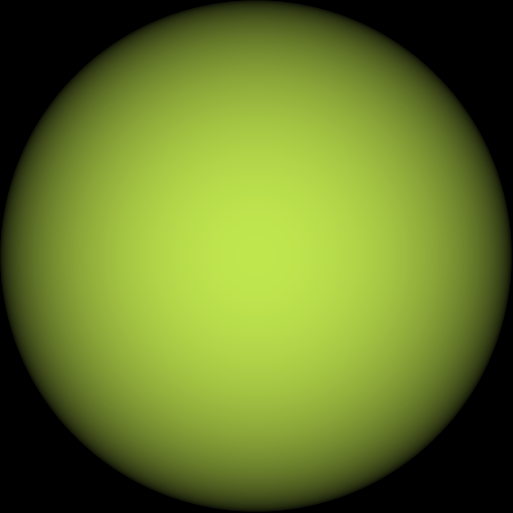}}  \hspace{-1.6mm}
  \subfigure[] {\label{synimg_ms}\includegraphics[width=24mm]{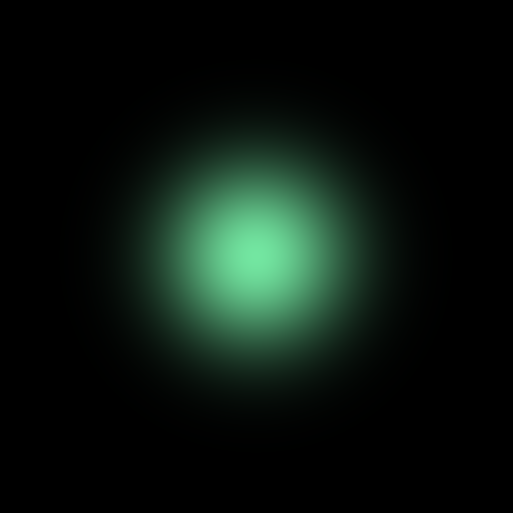}} \\
  \subfigure[] {\label{smallRatio}\includegraphics[height=32mm]{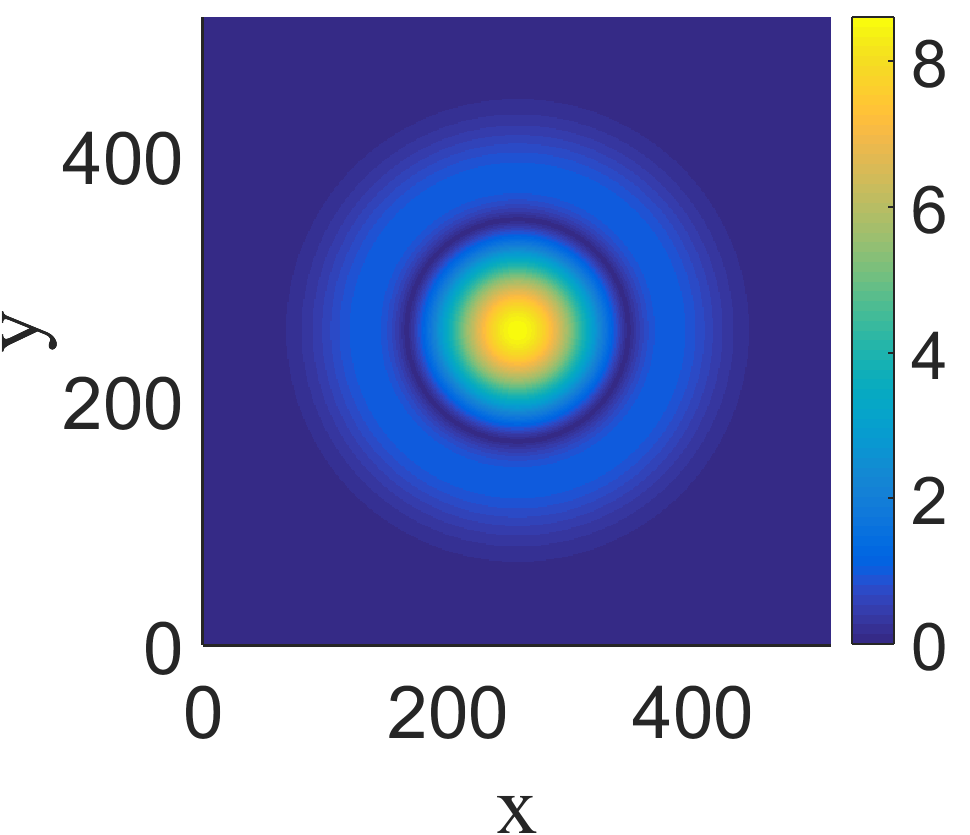}}\hspace{2mm}
  \subfigure[] {\label{largeRatio}\includegraphics[height=32mm]{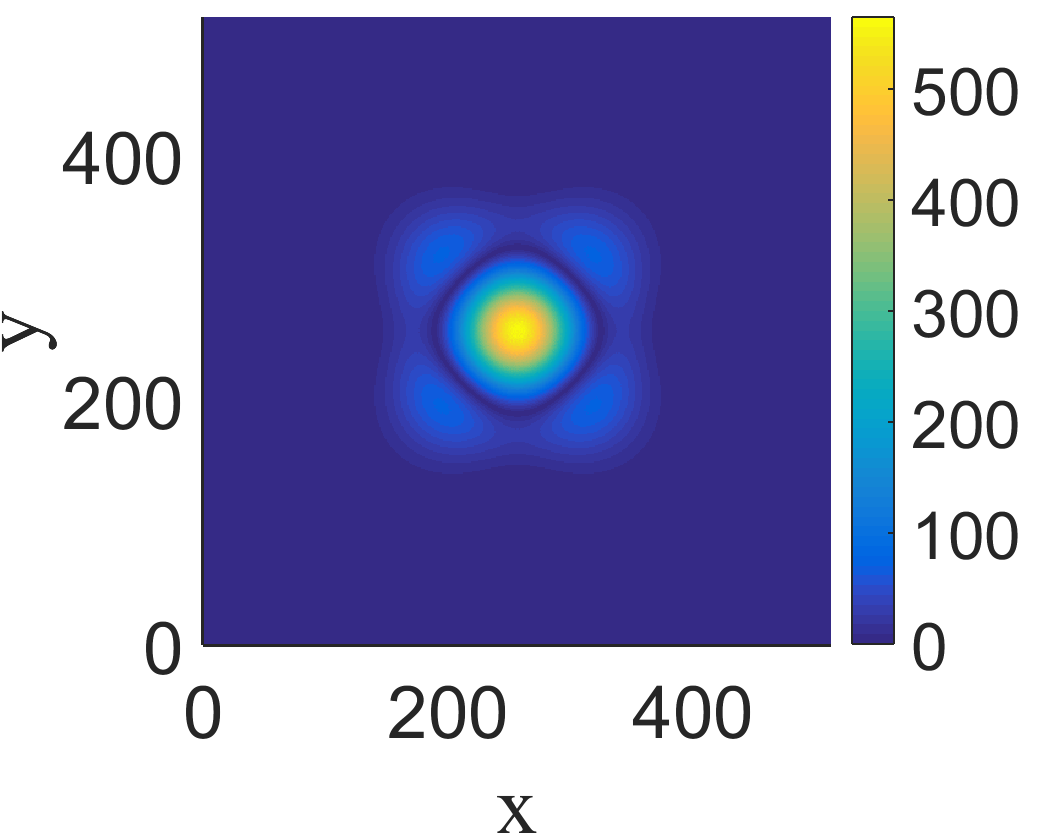}} \\
   \caption{(a)The synthetic image rendered
using Torrance-Sparrow reflection model \cite{torrance1967theory}. (b)The diffuse component of the synthetic image.
(c)The specular component of the synthetic image.
(d)The ratio $\frac{|\triangle m_s|}{|\triangle m_d|}$ in different locations after differentiating the image
with operator $\ve{f}_1$. (e)The ratio $\frac{|\triangle m_s|}{|\triangle m_d|}$ in different locations
after differentiating the image
with operator $\ve{f}_2$.}\label{different_ratio}
\end{figure}

\subsection{Derivative Color}
\label{theory}
To obtain the illuminant color based on specular components of the dichromatic reflection model, naively we consider a pair of highlight pixels
$\ve{I}(\ve{x})$, $\ve{I}(\ve{x}')$ with the same intrinsic object color $\ve{D}$. By differing
them, we obtain
\begin{equation}\label{naive}
\ve{I}(\ve{x})-\ve{I}(\ve{x}')=\triangle m_d \ve{D} + \triangle m_s\ve{S}.
\end{equation}
If the ratio $\frac{|\triangle m_s|}{|\triangle m_d|}$
is large enough to make
\begin{equation}\label{approximate}
\triangle m_d \ve{D} + \triangle m_s\ve{S} \approx \triangle m_s\ve{S},
\end{equation}
the illuminant color will be successfully extracted. We explore such a possibility from the physical standpoint.

In fact, according to the Lambert's Law \cite{lambert1760law}, the magnitude of diffuse reflection $m_d$ depends on the diffuse albedo $k_d$, intensity
of incident illuminant $L$, together with the angle $\psi$ between the illumination direction and the surface normal. It can be expressed
simply as
\begin{equation}\label{model_md}
m_d \sim k_d L \cos\psi.
\end{equation}
Usually, variations on $L$ and $\psi$ are small for local points in the process of
image formation. As a result, for a local region of uniform intrinsic object color, $m_d$ will also
vary slightly since they will share the same diffuse albedo $k_d$. In contrast, the magnitude of
specular reflection $m_s$ is quite sensitive to the geometric configurations between the illuminant, the object, and the camera,
which therefore makes it vary more widely in the local regions.
Based on the model on specularity that Torrance and Sparrow model presented in \cite{torrance1967theory},
 $m_s$ has the following characteristics as
\begin{equation}\label{model_ms}
m_s\sim \frac{FG}{\cos\theta}\exp(-\frac{\alpha^2}{2\phi^2}),
\end{equation}
in which $F$ is the Fresnel reflection and is derived from the Fresnel equation,
$G$ is the geometrical attenuation
factor, and $\phi$ is the surface roughness. In addition, $\theta$ is the angle between the surface normal
and the viewing direction, while $\alpha$ is the angle between the
surface normal and the bisector of the viewing direction and the
illumination direction.
This dependence of $m_s$ on the geometrical configurations
has also been exploited in IIC \cite{tan2004color}.

Since $m_s$ varies more sensitive than $m_d$,
one direct solution to obtain the expression $\triangle m_d\ve{D}+\triangle m_s\ve{S}$ similar to Eq. (\ref{naive})
with large ratio $\frac{|\triangle m_s|}{|\triangle m_d|}$
is to base on
the derivative structure of the image. Let $\ve{f}$ be an arbitrary differential operator, and $\ve{I}$ be an image satisfying
the dichromatic reflection model. The derivative structure of the image is then calculated as $\ve{J}=\ve{I}\otimes\ve{f}$.
If \ve{J}(\ve{x}) is obtained by convoluting \ve{I}(\ve{x}) with \ve{f} in a region of uniform intrinsic object
color \ve{D}, it can be simply expressed as
\begin{equation}\label{uniformJ}
 \ve{J}(\ve{x}) = \triangle m_d(\ve{x}) \ve{D}  + \triangle m_s(\ve{x}) \ve{S}.
\end{equation}
However, the differential operator \ve{f} should be carefully determined such that
the ratio $\frac{|\triangle m_s(\ve{x})|}{|\triangle m_d(\ve{x})|}$ obtained for \ve{J}(\ve{x}) in Eq. (\ref{uniformJ})
could be
as large as possible.
For example, we show typically a synthetic image $\ve{I}$ with uniform intrinsic object color \ve{D} in Fig. \ref{synimg}.
Fig. \ref{synimg_md} and Fig. \ref{synimg_ms} are respectively its diffuse and specular components.
By differentiating respectively the diffuse and specular components of image \ve{I} with the operator
\begin{equation}
\ve{f}_1 = \left[
             \begin{array}{rrr}
               \ \ 0& \ -1 & \ \ 0 \\
               \ -1 & \ \ 4 & \ -1 \\
               \ \ 0 & \ -1 & \ \ 0 \\
             \end{array}
           \right],
\end{equation}
we can readily obtain the ratio $\frac{|\triangle m_s|}{|\triangle m_d|}$ for each spatial point $\ve{x}$, which are altogether less than 10.
Fig. \ref{smallRatio} plots the result.
In contrast,
using the operator

\begin{equation}
\ve{f}_2 = \left(
             \begin{array}{rrrrrrr}
             \ \ \ 0 &\ \ \ \ 0 & \ \ \ 0 & \ \ -1
               & \ \ \ 0 &\ \ \ \ 0 & \ \ \ 0  \\
               \ \ \ 0 &\ \ \ \ 0 & \ \ \ 0 & \ \ \ 6
               & \ \ \ 0 &\ \ \ \ 0 & \ \ \ 0  \\
               \ \ \ 0 &\ \ \ \ 0 & \ \ \ 0 & \ -15
               & \ \ \ 0 &\ \ \ \ 0 & \ \ \ 0  \\
               \ \ -1 &\ \ \ \ 6 & \ -15 & \ \ 40
                &\ -15 &\ \ \ \ 6 & \ \ -1 \\
               \ \ \ 0 &\ \ \ \ 0 & \ \ \ 0 & \ -15
               & \ \ \ 0 &\ \ \ \ 0 & \ \ \ 0  \\
               \ \ \ 0 &\ \ \ \ 0 & \ \ \ 0 & \ \ \ 6
               & \ \ \ 0 &\ \ \ \ 0 & \ \ \ 0  \\
               \ \ \ 0 &\ \ \ \ 0 & \ \ \ 0 & \ \ -1
               & \ \ \ 0 &\ \ \ \ 0 & \ \ \ 0 \\
             \end{array}
           \right),
\end{equation}
the ratios we achieve can be greater than 500.
Fig. \ref{largeRatio} plots the result.
It can be seen that evidently, the operator $\ve{f}_2$ produces larger ratios than $\ve{f}_1$.
In addition, for both $\ve{f}_1$ and $\ve{f}_2$,
locations containing higher specular components are basically with larger ratios.
Note in this experiment,
we perform differentiations along the
horizontal and vertical directions for their simplicities.

\begin{figure}[!t]
  \centering
\hspace{1mm}
  \subfigure[] {\label{sphere_multi_1}\includegraphics[width=28mm]{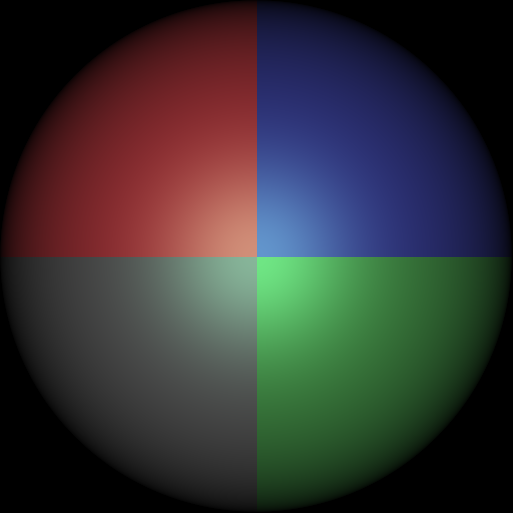}}  \hspace{12mm}
  \subfigure[] {\label{sphere_multi_2}\includegraphics[width=28mm]{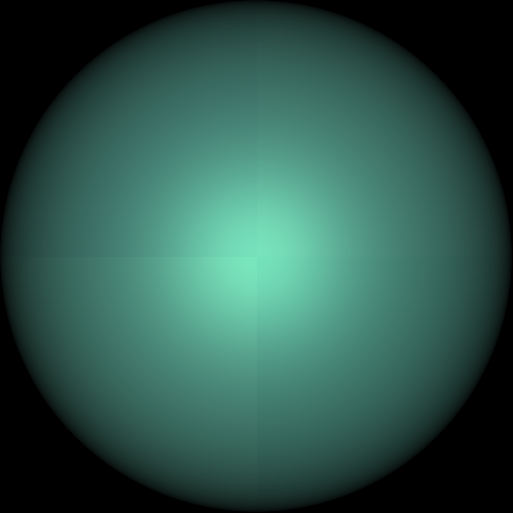}}
   \subfigure[] {\label{ratio_sphere_multi_f21}\includegraphics[width=41mm]{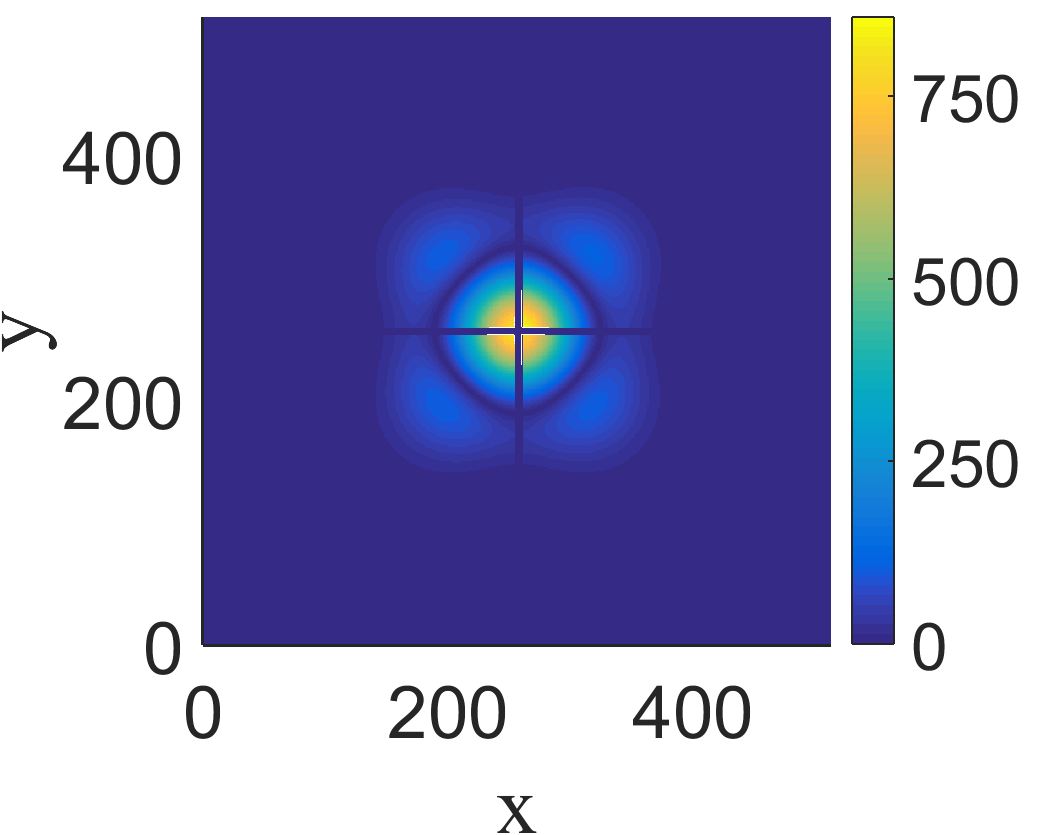}}
  \subfigure[] {\label{ratio_sphere_multi_f22}\includegraphics[width=41mm]{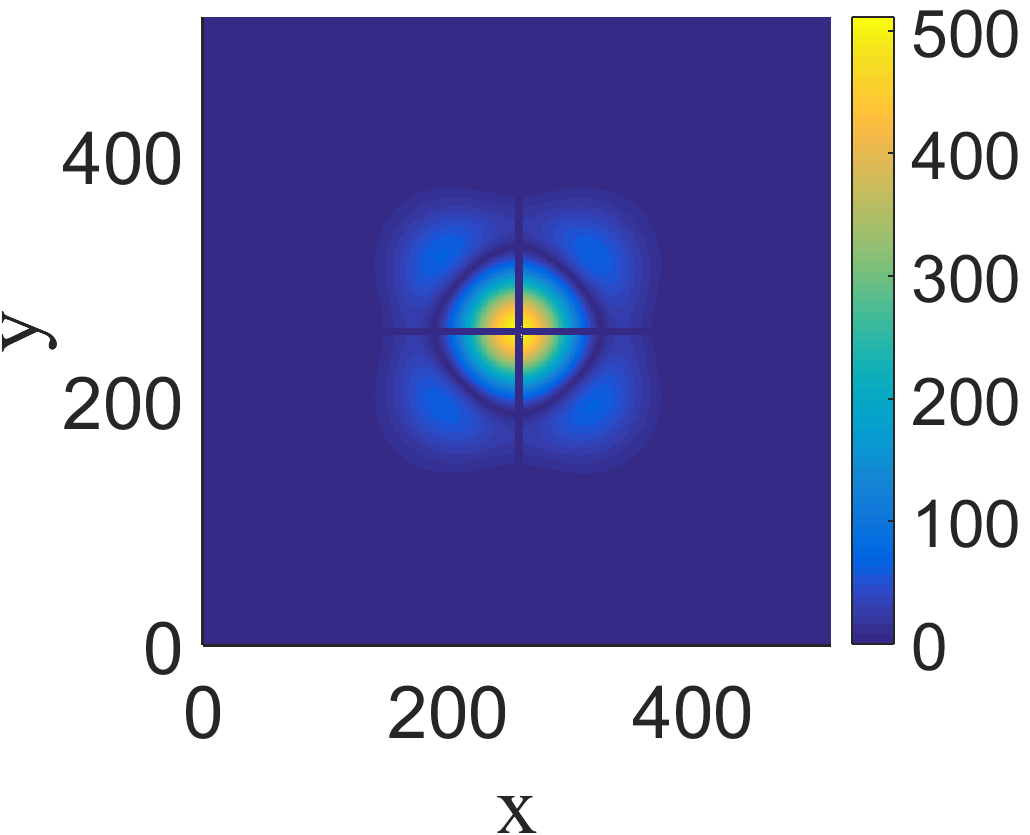}} \\
   \caption{(a)The synthetic image with distinctive intrinsic object colors. (b)
The synthetic image with similar intrinsic object colors.
(c)The ratio $\frac{|\triangle m_s|}{|\triangle m_d|}$ in different locations after differentiating the image in (a)
with operator $\ve{f}_2$. (d)The ratio $\frac{|\triangle m_s|}{|\triangle m_d|}$ in different locations
after differentiating the image in (b)
with operator $\ve{f}_2$.}\label{multiple_color_ratio}
\end{figure}
Beyond uniform regions, the ratio for \ve{J}(\ve{x}) which is obtained by convoluting \ve{I}(\ve{x}) with \ve{f} in a region of non-uniform intrinsic object colors
is explored as well.
In fact, each $\ve{J}(\ve{x})$, either obtained in uniform regions or non-uniform regions, can be altogether represented as
\begin{equation}\label{arbitraryJ}
 \ve{J}(\ve{x}) = \triangle m_d(\ve{x}) \ve{D}'(\ve{x})  + \triangle m_s(\ve{x}) \ve{S},
\end{equation}
where $\triangle m_d(\ve{x}),\triangle m_s(\ve{x})\in(-\infty,\infty)$, and $\ve{D}'(\ve{x})$ is an arbitrary vector with
$\sum_k D_k'(\ve{x})=1$.
The expression shown in Eq. (\ref{uniformJ}) is just a special case of Eq. (\ref{arbitraryJ}), with $\ve{D}'(\ve{x})=\ve{D}$.
Usually, for non-uniform regions, $|\triangle m_d|$ is quite large, which as a result leads the ratio $\frac{|\triangle m_s|}{|\triangle m_d|}$ to be small. We show two such examples in Fig. \ref{multiple_color_ratio}.
Fig. \ref{sphere_multi_1} is a synthetic image with four distinctive intrinsic object colors. Fig. \ref{sphere_multi_2}
is a synthetic image with four similar intrinsic object colors. Figs. \ref{ratio_sphere_multi_f21}, \ref{ratio_sphere_multi_f22}
are respectively their ratios in different locations after differentiated by operator $\ve{f}_2$. It can be seen that
in a non-uniform region, even when the intrinsic object colors are close to each other, the ratio achieved is still quite small.
To this end, we introduce the definition on the derivative color as follows. Each
\ve{J}(\ve{x}) in the derivative structure \ve{J} of an image \ve{I}, is denoted as a \textbf{derivative color} if $\frac{J_k(\ve{x})}{\sum_k J_k(\ve{x})}\in(0,1), k\in\{r,g,b\}$. Let
\begin{equation}\label{DCchrom}
     c_r(\ve{x}) = \frac{J_r(\ve{x})}{\sum_k J_k(\ve{x})},
     c_g(\ve{x}) = \frac{J_g(\ve{x})}{\sum_k J_k(\ve{x})},
     c_b(\ve{x}) = \frac{J_b(\ve{x})}{\sum_k J_k(\ve{x})}.
     \notag
\end{equation}
We refer $c_r(\ve{x})$, $c_g(\ve{x})$, $c_b(\ve{x})$ as the red, green and blue chromaticities of the derivative color \ve{J}(\ve{x})
respectively. Since uniform regions generally produces large ratio $\frac{|\triangle m_s(\ve{x})|}{|\triangle m_d(\ve{x})|}$ for \ve{J}(\ve{x}) than non-uniform regions,
we focus on estimating the illuminant color using derivative colors obtained from uniform regions.

\begin{figure}[!t]
  \centering
\subfigure[] {\label{ratio_error}\includegraphics[width=41mm]{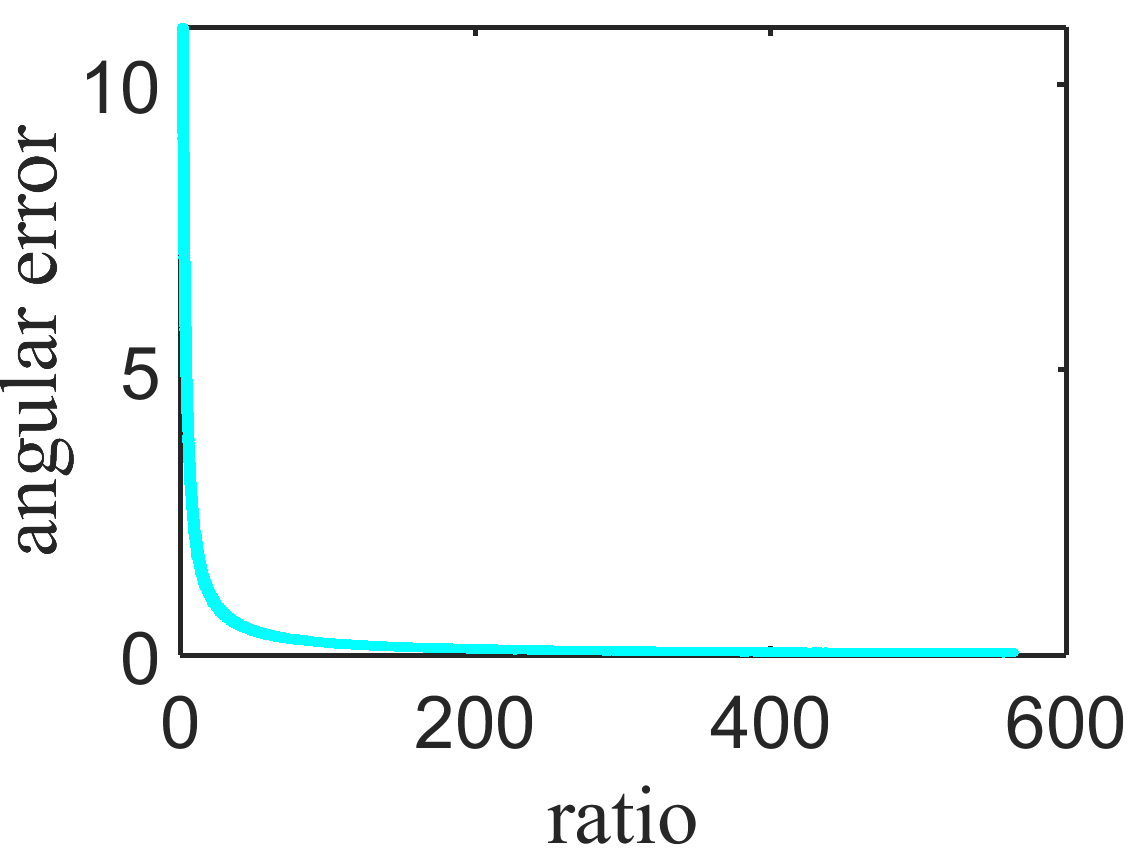}}
\subfigure[] {\label{ratio_rg}\includegraphics[width=41mm]{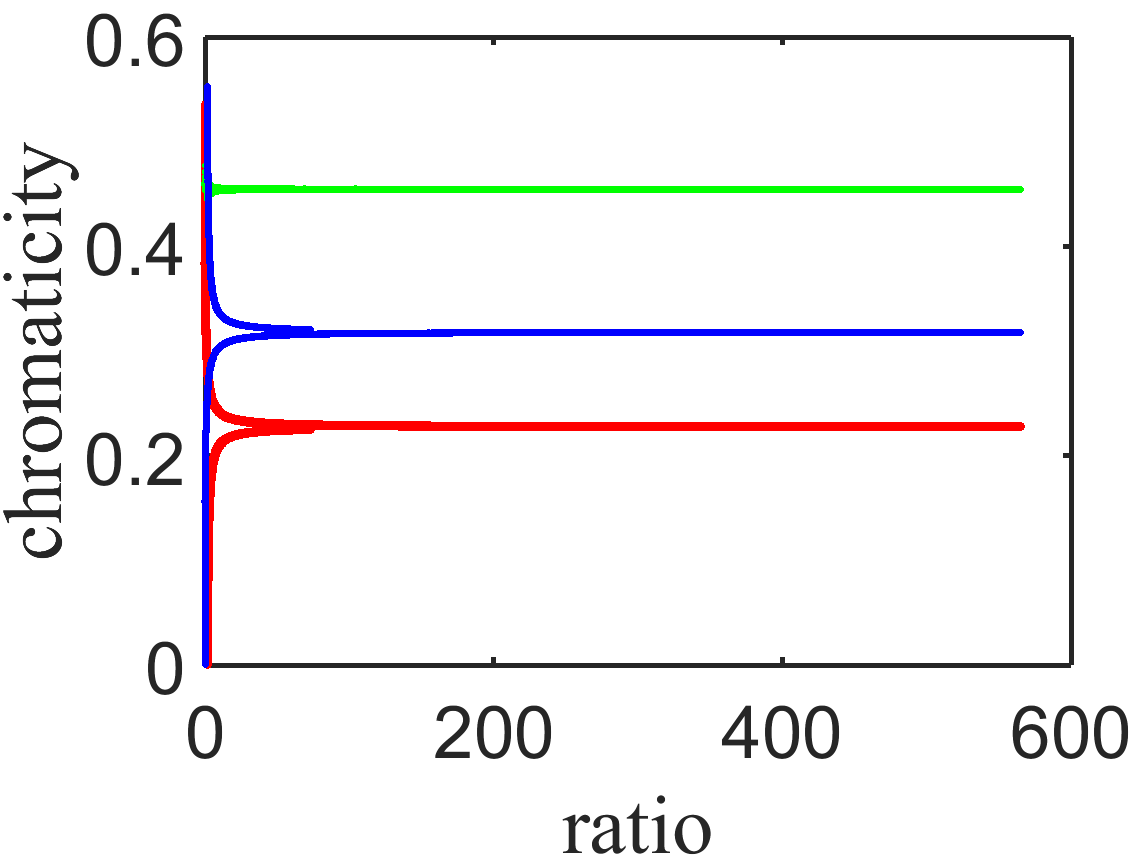}}
\caption{(a)The variation of the angular error according to the ratio $\frac{|\triangle m_s|}{|\triangle m_d|}$.
(b)The variations of chromaticities of the estimated illuminant color \ve{J}(\ve{x}) according to the ratio.
We plot the red, green and blue chromaticities correspondingly with red, green and blue dots.
}\label{estimation_accuracy}
\end{figure}

Undoubtedly, 
the larger the ratio $\frac{|\triangle m_s(\ve{x})|}{|\triangle m_d(\ve{x})|}$ is,
the better the derivative color $\ve{J}(\ve{x})$ will approximate to $\triangle m_s\ve{S}$.
Taking each derivative color $\ve{J}(\ve{x})$ from the image shown in Fig. \ref{synimg}
as an estimate of the illuminant color, we track the variation of the estimation accuracy according
to the ratio with experiment.
In addition, the angular error is adopted to measure our estimation accuracy,
 which is recently referred as the recovery error in \cite{finlayson2014reproduction}. In particular, it is defined as \cite{hordley2006reevaluation}
\begin{equation}\label{error}
err_{angle}(\ve{S},\ve{S}_{est})=cos^{-1}\left(\frac{\ve{S}\cdot\ve{S}_{est}}{\|\ve{S}\|\|\ve{S}_{est}\|}\right),
\end{equation}
where \ve{S} is the ground truth illuminant color, and $\ve{S}_{est}$ is the estimated illuminant color.
Specifically, here we have $\ve{S}_{est} = \ve{J}(\ve{x})$.
Since larger ratio means better estimation,
it will therefore contribute to smaller angular error.
Fig. \ref{ratio_error} shows the experimental result, which consists well with our speculation.
Besides, it can be seen that when the ratio is larger than 30 around,
the angular error is close to 0 and varies quite slightly,
which is caused by the small variations
on the chromaticities of
the derivative color \ve{J}(\ve{x}).
We show variations on the red, green and blue chromaticities of $\ve{J}(\ve{x})$ according to the ratio in Fig. \ref{ratio_rg}. It can be seen that they altogether keep nearly constant when the ratio is larger than 30.

We have validated the theoretical feasibility of estimating the illuminant color with derivative colors extracted from uniform highlight regions. As to pixels in achromatic regions, which can be generally expressed as
 \begin{equation}\label{BRDF_achromatic}
 \begin{aligned}
\ve{I}(\ve{x})&= (m_d(\ve{x})+m_s(\ve{x}))\ve{S},\\
\end{aligned}
\end{equation}
the ratio $\frac{|\triangle m_s|}{|\triangle m_d|}$ of each derivative color $\ve{J}(\ve{x})$ can be actually considered as $\infty$. That is, derivative color from achromatic regions are identical to the illuminant color. Specifically, such successful extraction in achromatic regions are due to either diffuse shadow or shading (variations on $m_d$), or highlight (variations on $m_s$). Therefore, using derivative colors, we can achieve color constancy from both uniform highlight regions and achromatic regions.

\subsection{Illuminant Estimation}
\label{practice}
\begin{figure*}
\centering
 \subfigure {\includegraphics[width=40mm]{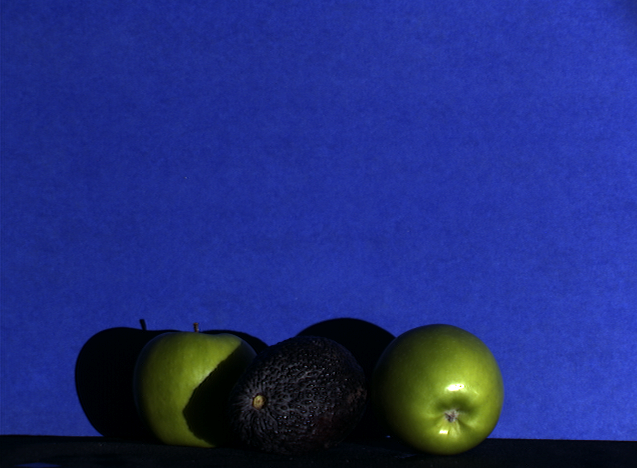}}
 \subfigure {\includegraphics[width=40mm]{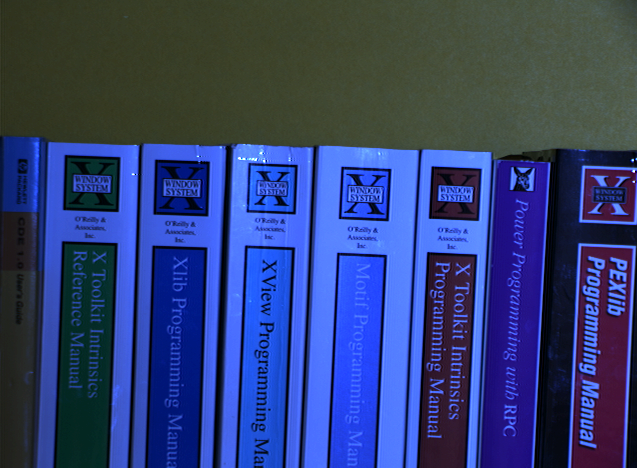}}\hspace{0.1mm}
\subfigure {\includegraphics[width=40mm]{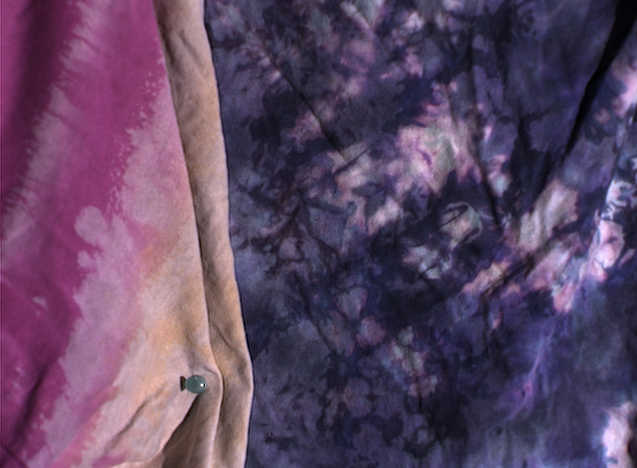}}\hspace{0.1mm}
\subfigure {\includegraphics[width=40mm]{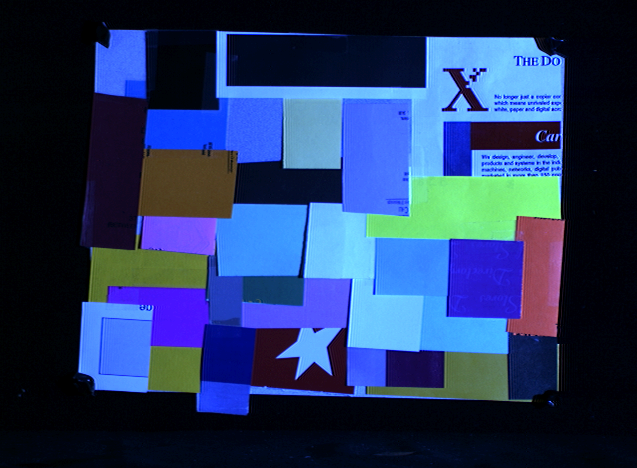}} \\
 \subfigure {\includegraphics[width=41mm]{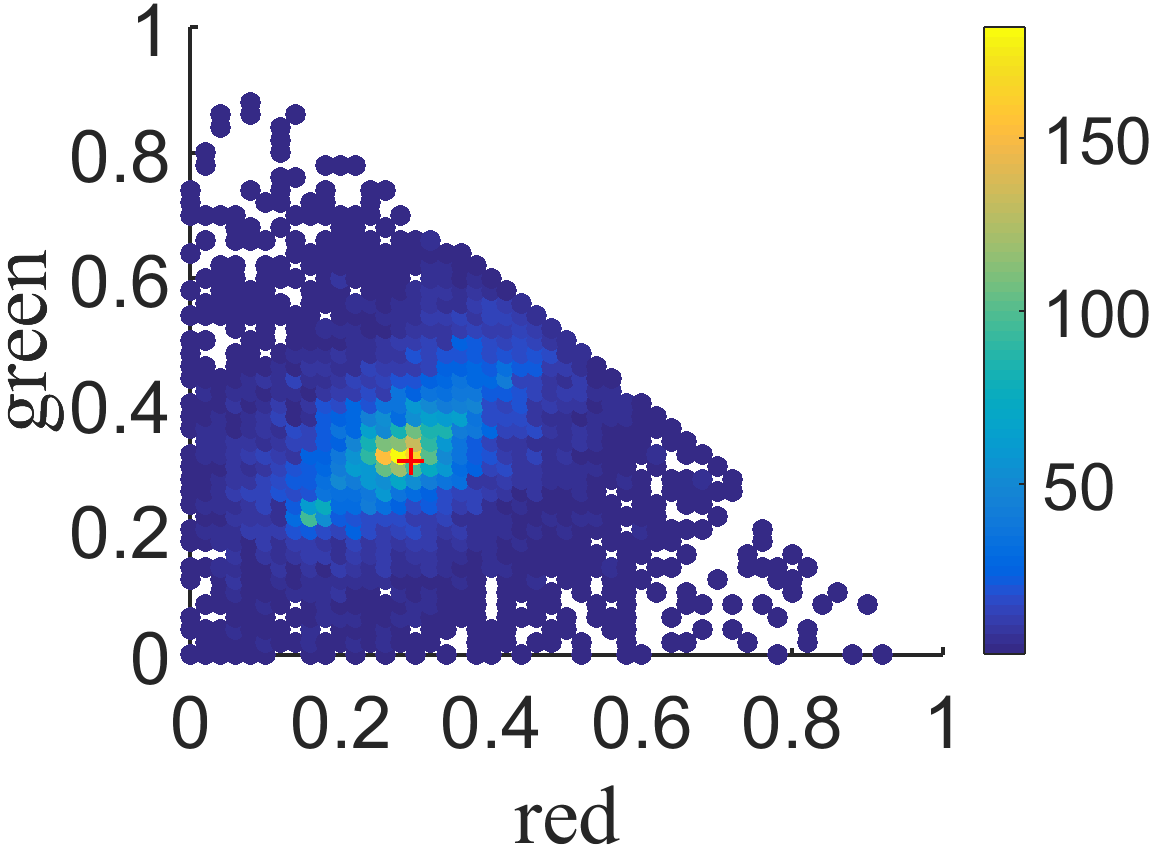}}
 \subfigure {\includegraphics[width=41mm]{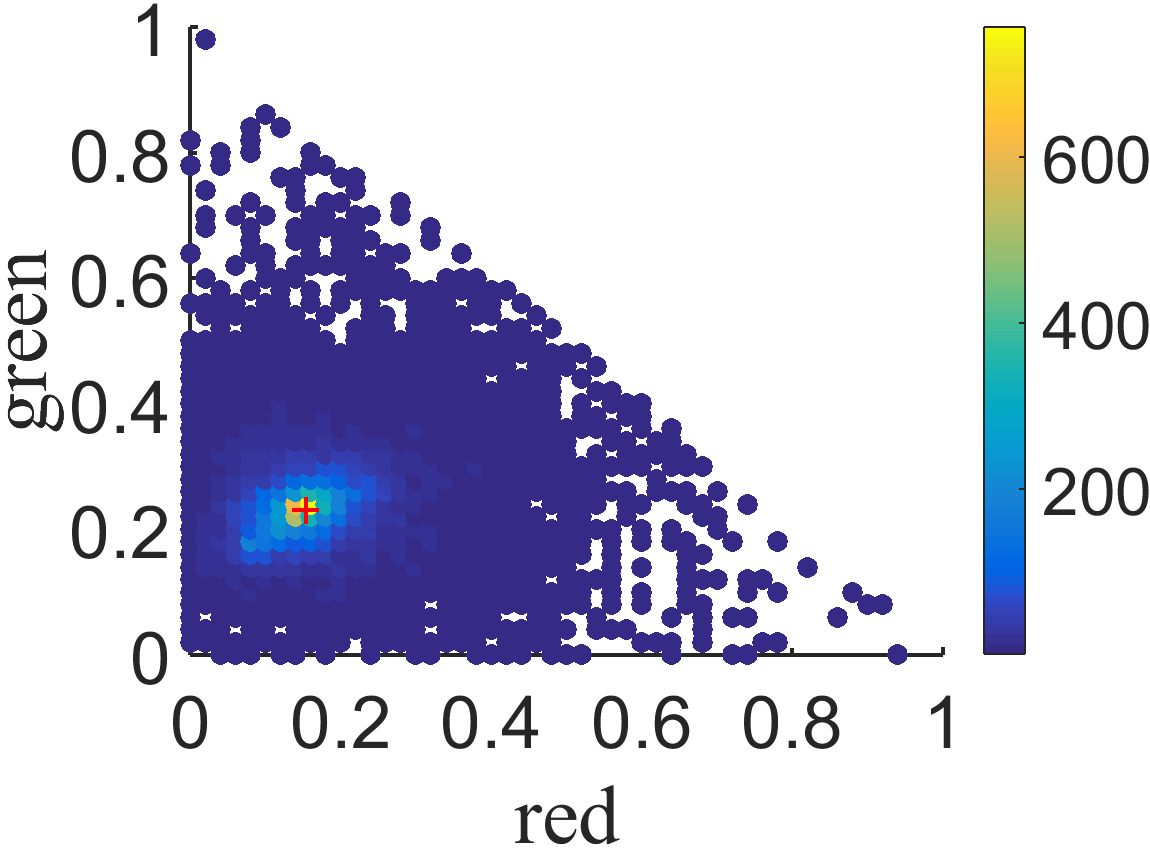}}
\subfigure {\includegraphics[width=41mm]{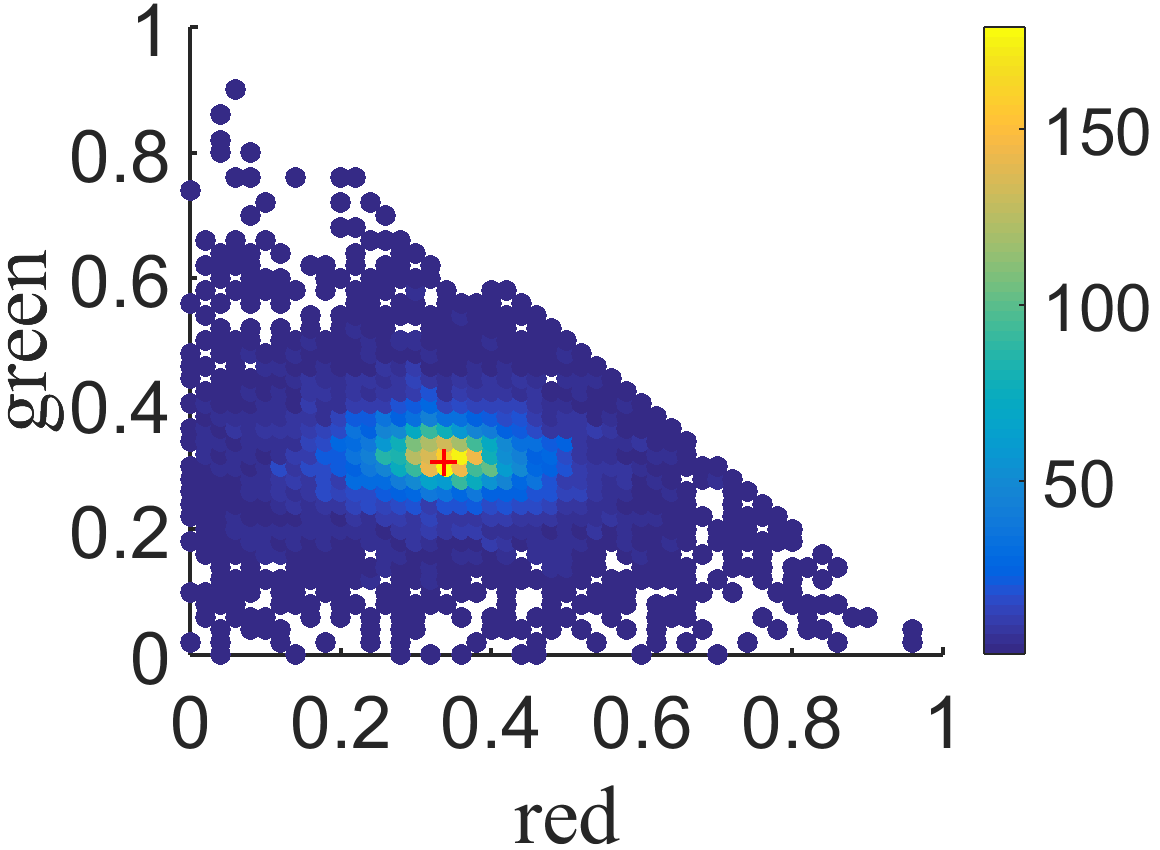}}
\subfigure {\includegraphics[width=41mm]{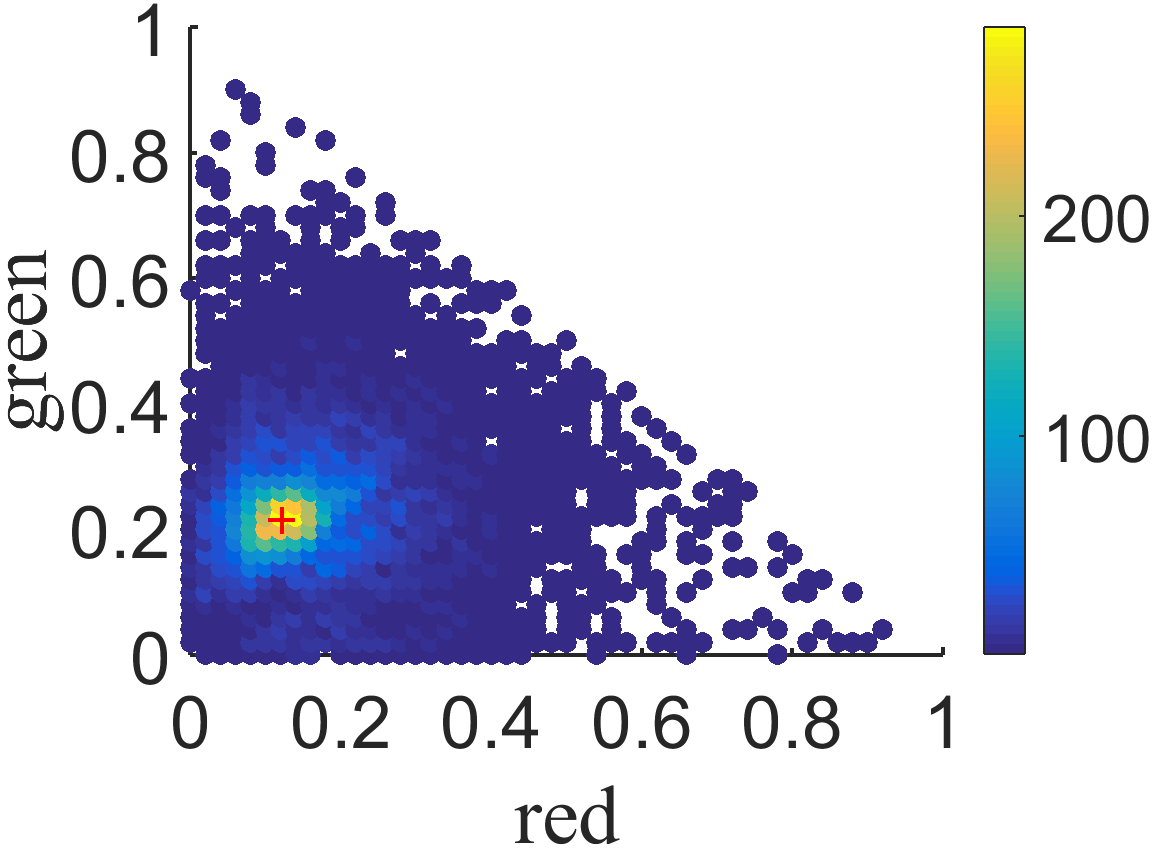}}\\
 \subfigure {\includegraphics[width=40mm]{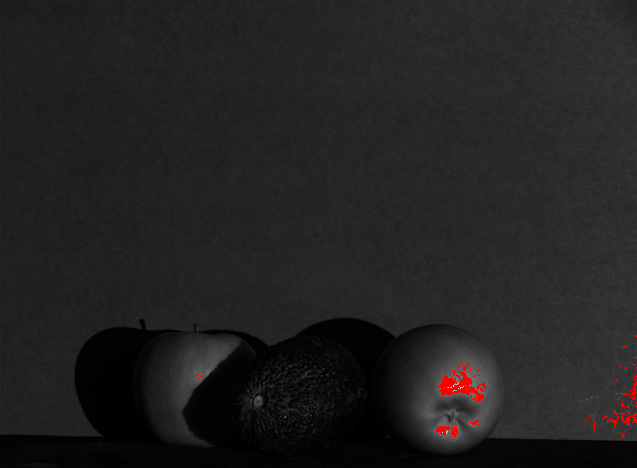}}
 \subfigure {\includegraphics[width=40mm]{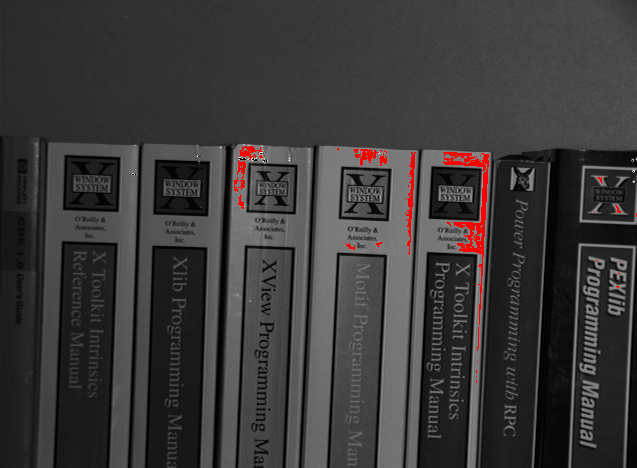}}\hspace{0.1mm}
\subfigure {\includegraphics[width=40mm]{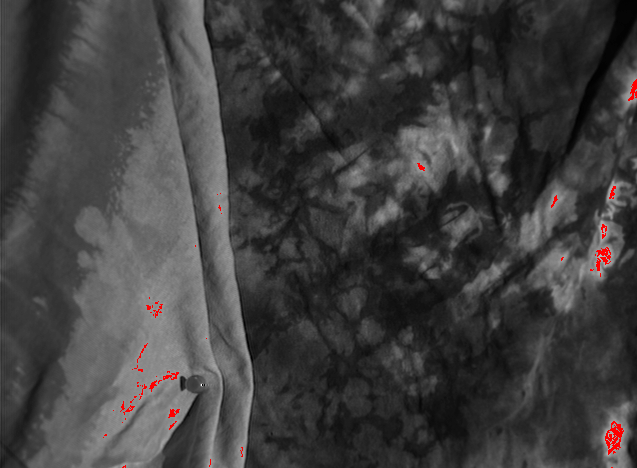}}\hspace{0.1mm}
\subfigure {\includegraphics[width=40mm]{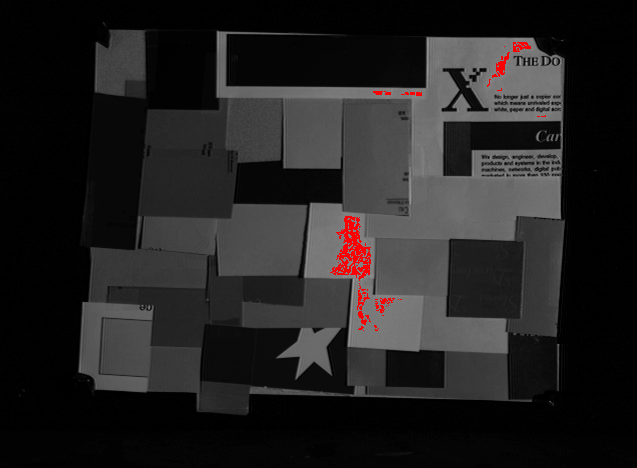}} \\
  \caption{Example on distributions of $\{\ve{z}_i\}_{i=1}^{n}$. The first row shows the original images. The
 second row shows their distributions in the rg chromaticity space correspondingly.
  We plot points in $\{\ve{z}_i\}_{i=1}^{n}$ with color coded dots such that values of these colors explicitly
represent frequencies of the points. In addition, the ground truth is plotted with red cross.
  It can be seen that the majority of the data are densely distributed around the ground truth.
And the third row shows the gray-scale of the original images, on which
 the pixels whose derivative colors are with frequencies larger than 50 are labelled as red.
It can be seen that they generally locate in the uniform regions.
}\label{kernel}
\end{figure*}
In reality, it is readily achievable to extract derivative colors identical to the illuminant color from
achromatic regions. Yet, to achieve color constancy
using the derivative colors from uniform highlight regions of real images,
two issues should
be addressed as effectively as possible. The first one is that in order to
obtain large enough ratio
$\frac{|\triangle m_s|}{|\triangle m_d|}$,
what kind of
differential operators can be used to extract the derivative colors.
Besides,
since it is usually impossible to obtain the diffuse and specular components of an image separately,
the ratio $\frac{|\triangle m_s|}{|\triangle m_d|}$ therefore becomes unavailable for each location.
As a result, we cannot determine which derivative color is a better estimate of the illuminant color.
 Thus, the second issue is that what kind of pixels generally
produce larger ratios and can be used to estimate the illuminant color better.

As a matter of fact, due to the universal existence of image noise, quantization error, and textured surfaces et al.,
the differential operator should be determined with efforts.
However, based on the recent success of the gaussian derivative filters in a number of
works (e.g., \cite{van2007edge, gijsenij2010generalized, chakrabarti2012color, finlayson2013corrected}), we found that a combination of the second-order gaussian functions $\ve{f}_{xx}$,
$\ve{f}_{yy}$, $\ve{f}_{xy}$ works well in extracting the derivative colors from highlight regions as well as achromatic regions.
Particularly, the differential operators take the following expressions,
\begin{equation}\label{2ndOperator}
\left\{
  \begin{array}{ll}
    \ve{f}_{xx} = \big(1-\frac{x^2}{\sigma^2}\big)\exp\big(-\frac{x^2}{2\sigma^2}\big), \\
    \ve{f}_{yy} = \big(1-\frac{y^2}{\sigma^2}\big)\exp\big(-\frac{y^2}{2\sigma^2}\big), \\
    \ve{f}_{xy} = xy\exp\big(-\frac{x^2+y^2}{2\sigma^2}\big),
  \end{array}
\right.
\end{equation}
where the constants in these expressions are omitted for convenience since they have no any influence on the illuminant estimation.
The parameter $\sigma$ controls the scale of the convolution. And we expect $\sigma$
to be small so that more derivative colors are extracted from uniform regions.
The derivative structures of an image obtained from these differential operators are
represented as
$\ve{J}_{xx}=\ve{I}\otimes \ve{f}_{xx}$, $\ve{J}_{yy}=\ve{I}\otimes \ve{f}_{yy}$, and
$\ve{J}_{xy}=\ve{I}\otimes \ve{f}_{xy}$ respectively.
We denote the set of locations for the
derivative colors in these derivative structures as $\mathcal{L}_{xx}$, $\mathcal{L}_{yy}$, $\mathcal{L}_{xy}$
correspondingly.

Sufficiently, for highlight pixels, if $|\triangle m_s|$ is large and $|\triangle m_d|$ is small,
the ratio $\frac{|\triangle m_s|}{|\triangle m_d|}$ will be large. In order
to obtain large $|\triangle m_s|$, highlight regions with strong specular components
are good choices.
However, to the best of our knowledge, the detection of highlight
is still a challenging and open problem in computer vision at the current stage, even though some solutions
are suggested (e.g., \cite{delpozo2007detecting,angelopoulou2007specular,yilmaz2014detection}).
In addition, without guidance of the illuminant color, it is also impossible to identify achromatic regions.
Due to this fact,
we detect the expected regions in a compromised way, which is similar to
the way adopted in \cite{tan2004color, joze2012role, drew2014zeta}.
Firstly, sort the intensity of all pixels after the saturated pixels are clipped.
 Secondly, the $\tau\%$ pixels with higher intensity are labelled to build a binary mask for the original image.
Finally,
erode the binary mask until no more than $\eta\%$ pixels are left.
The reasons facilitating our detection mechanism are two-fold.
On the one hand, highlight regions with high specular components are usually much brighter.
In addition, white patches, which are achromatic, generally have higher intensity as well.
Considering the difficulty of detecting achromatic and highlight regions strictly,
using bright regions as a replacement is a reasonable trade-off in reality.
On the other hand, since the small regions in the binary mask have a high
probability to be noise and unreliable, we add the erosion operation to remove them,
which indeed stabilizes 
 the estimation accuracy as shown in the experimental part later.
The set of locations for all pixels in the finally selected regions is denoted as $\mathcal{L}^h$.

We then represent the selected derivative colors
as
\begin{equation}\label{selected_derivative_color}
\mathcal{J}^h=\mathcal{J}_{xx}^h\bigcup\mathcal{J}_{yy}^h\bigcup\mathcal{J}_{xy}^h,
\end{equation}
in which
\begin{equation}
\left\{
  \begin{array}{ll}
    \mathcal{J}_{xx}^h=\{\ve{J}_{xx}(\ve{x})|\ve{J}_{xx}=\ve{I}\otimes \ve{f}_{xx},\ve{x}\in\mathcal{L}_{xx}\bigcap\mathcal{L}^h\}, \\
    \mathcal{J}_{yy}^h=\{\ve{J}_{yy}(\ve{x})|\ve{J}_{yy}=\ve{I}\otimes \ve{f}_{yy},\ve{x}\in\mathcal{L}_{yy}\bigcap\mathcal{L}^h\}, \\
    \mathcal{J}_{xy}^h=\{\ve{J}_{xy}(\ve{x})|\ve{J}_{xy}=\ve{I}\otimes \ve{f}_{xy},\ve{x}\in\mathcal{L}_{xy}\bigcap\mathcal{L}^h\}.
  \end{array}
\right.
\notag
\end{equation}
For the selected derivative colors in achromatic regions of $\mathcal{J}^h$, perfect illuminant color is obtained.
For the selected derivative colors in highlight regions of $\mathcal{J}^h$, large $|\triangle m_s|$ is obtained.
As to $|\triangle m_d|$, it is impossible to ensure it to be small for all pixels in $\mathcal{L}^h$.
The reason is that $|\triangle m_d|$ is usually small in uniform region and large in non-uniform region,
while real images generally suffer from the influence of varying intrinsic object colors.
Therefore,
it is impossible to obtain large ratio $\frac{|\triangle m_s|}{|\triangle m_d|}$ for all pixels in
$\mathcal{L}^h$. However, derivative colors from
the uniform highlight regions are generally with large ratios, which therefore makes them close to the illuminant color.
Projecting the selected derivative colors in $\mathcal{J}^h$ into the rg chromaticity
space, we observe that despite with outliers, the majority of the data distribute densely around the ground truth.
Fig. \ref{kernel} shows several examples. 
The first row is the original images. The
 second row is their distributions in the rg chromaticity space correspondingly.
    It can be seen that the majority of the data are densely distributed around the ground truth.
And the third row shows pixels whose derivative colors are with frequencies larger than 50.
It can be seen that they generally locate in the uniform regions.

With some abuse of notation, the data set projected into the rg chromaticity space is represented
as $\{\ve{z}_i\}_{i=1}^{n}$,
and $n=|\mathcal{J}^h|$. $\ve{z}_i$ is computed as
$\ve{z}_i = [c_r(\ve{x}_i)\ c_g(\ve{x}_i)]^T$. And $c_r(\ve{x}_i)$, $c_g(\ve{x}_i)$
are respectively the red and green chromaticities of the derivative color $\ve{J}(\ve{x}_i)\in\mathcal{J}^h$.
We then present to take the point with the maximum density in the rg chromaticity space as an estimate
of the illuminant chromaticity.
And the density of each point $\ve{z}$ is computed using the following function:
\begin{equation}\label{density}
p(\ve{z})=\frac{1}{n}\sum_{i=1}^n k(\ve{z}-\ve{z}_i),
\end{equation}
where $k(\cdot)$ is the Parzen kernel estimator (which has the properties of non-negative and integrating to 1).
Its typical form is a Gaussian:
$k(\ve{z}-\ve{z}_i)=\exp(-\frac{1}{2h^2}\|\ve{z}-\ve{z}_i\|^2)$ \cite{duda1999pattern}, with $h$ controlling
the smoothness of the kernel function on the data.
Then, the estimated illuminant chromaticity is taken as
\begin{equation}\label{result}
\ve{z}_{est}=\arg\max_{\ve{z}_i} p(\ve{z}_i).
\end{equation}
An estimate of the illuminant color can be readily obtained from
$\ve{z}_{est}=[c_r\ c_g]^T$, which
equals
\begin{equation}\label{estimateIlluminant}
\ve{S}_{est} = [c_r\ c_g\ 1-c_r-c_g]^T.
\end{equation}

Since the proposed approach is based on a key exploitation of the derivative colors,
we abbreviate our algorithm as the DCs algorithm and summarize it in Algorithm \ref{alg:Estimate}.
\begin{algorithm}[!htb]
\caption{}
\label{alg:Estimate}
\begin{tabular}{ll}
\hspace{-3.3mm}
  \textbf{Objective:} & \hspace{-2mm}Estimate the illuminant color $\ve{S}_{est}$ of a color \\
  {} & \hspace{-2mm}image $\ve{I}$ ; \\
\end{tabular}
\textbf{Steps:}
\begin{algorithmic}[1]
\STATE Clip the saturated pixels and select $\tau\%$ pixels with higher intensity to build a binary mask;
\STATE Erode the binary mask to remove small segments until no more than $\eta\%$ pixels are left;
\STATE Extract the derivative colors from the selected locations $\mathcal{L}^h$, using the differential operators $\ve{f}_{xx}$, $\ve{f}_{yy}$ and $\ve{f}_{xy}$;
\STATE Compute the probability for each data in the rg chromaticity space with kernel density estimation;
\STATE Estimate the data with the maximum density, $\ve{z}_{est}$;
\STATE Compute the illuminant color $\ve{S}_{est}$ based on $\ve{z}_{est}$.
\end{algorithmic}
\end{algorithm}

\section{Experimental Evaluation}

We evaluate the proposed approach on three standard databases:
the SFU laboratory database \cite{barnard2002comparison}, the Gehler-Shi database \cite{shi2010data},
and the SFU HDR database \cite{funt2010rehabilitation}.
The error between the ground truth \ve{S} and the estimated illuminant color $\ve{S}_{est}$ is
computed according to Eq. (\ref{error}).
In addition, besides the median and mean angular errors, we report
 the trimean, the \emph{Best 25 percent} or the \emph{Worst 25 percent} selectively on each database in order to compare the performance of different approaches better. Among these measures, the
median indicates performance of the method
on the majority of the images, while the trimean also provides an
indication on the extreme values of the distribution.  The best 25 percent and worst 25 percent errors are robust measures which refer to the mean of the 25 percent lowest and highest
error values respectively. Further, to summarize the performance of different algorithms
with more insight, a \emph{sign test} \cite{hordley2006scene} is conducted between each pair of them
over every database.
 This sign test
determines whether one algorithm tends to have lower errors
compared to another by using significance testing to reject the
null hypothesis that the medians of the error distributions of two algorithms are the same.
Commonly, the confidence level for accepting the null hypothesis is chosen as 95\% (e.g., \cite{gijsenij2011computational,chakrabarti2012color,gao2015color}).

According to our categorization in the introduction, we consider the existing algorithms in our comparison
mainly from two categories: (1) physics-based methods: using natural illuminant colors as constraints (NICs) \cite{finlayson2001convex},
inverse-intensity chromaticity space (IIC) \cite{tan2004color}, and Zeta-image \cite{drew2014zeta}; (2) statistics-based methods: gray-world (GW) \cite{buchsbaum1980spatial},
white-patch (WP) \cite{land1977retinex},
 shades of gray (SoG) \cite{finlayson2004shades}, general gray-world (GG),
gamut mapping (GM(pixel)) \cite{forsyth1990novel}, Bayesian \cite{gehler2008bayesian}, neural network (NN) \cite{cardei2002estimating}, support vector regression (SVR)
\cite{funt2004estimating}, gray-edge (GE1 and GE2) \cite{van2007edge}, natural image statistics (NIS) \cite{gijsenij2011color}, generalized gamut mapping (GM($n$jet)) \cite{gijsenij2010generalized}, weighted gray-edge (WGE) \cite{gijsenij2012improving},
spatio-spectral statistics (SS) \cite{chakrabarti2012color},  PCA on dark and bright pixels (DBPCA) \cite{cheng2014illuminant},
corrected-moment (CM) \cite{finlayson2013corrected}, and multi-cue (MC) \cite{li2015multi}.
The biological-based method double-opponency (DOCC(sum and max)) \cite{gao2015color} is also included.
The error distributions of most approaches are directly available from websites \cite{cc_website}, \cite{li_website_all}.
The results of IIC, WGE, DBPCA, and GM($n$jet) are obtained by executing the codes published by the authors when the direct results are
unavailable on a database.
The results of MC are from website \cite{li_website_mc}. And those of DOCC are from website
 \cite{docc_website}. Please note that our results of DOCC on the latest Gehler-Shi database are obtained from the authors via
 private communication.

 \begin{table*}[!t]
 \caption{Performance on the SFU laboratory database.}\label{sfulab}
  \centering
  \small
  \begin{tabular}{|l|c|c|c|c|c|}
    \hline
Method & Median & Mean & Tri-mean & Best-25\% & Worst-25\% \\

    \hline
    \hline
    DN  & 15.60$^\circ$ & 17.27$^\circ$ & 16.56$^\circ$ & 3.63$^\circ$  &  32.49$^\circ$ \\
    \hline
GW & $7.00^\circ$ & $9.78^\circ$ & $7.59^\circ$ &  0.91$^\circ$ &  23.45$^\circ$  \\
    WP & $6.48^\circ$ & $9.09^\circ$ & $7.46^\circ$ & 1.86$^\circ$&20.97$^\circ$  \\
    SoG & $3.74^\circ$ & $6.39^\circ$ & $4.60^\circ$ & 0.60$^\circ$ &16.49$^\circ$ \\
SS & 3.45$^\circ$ & 5.63$^\circ$ & 4.33$^\circ$ & 1.24$^\circ$ &12.90$^\circ$ \\
     GG & $3.32^\circ$ & $5.41^\circ$ & $3.78^\circ$ & 0.50$^\circ$  &13.75$^\circ$  \\
GE1 & 3.18$^\circ$ & 5.58$^\circ$ & 3.75$^\circ$ &  1.07$^\circ$ &14.05$^\circ$ \\
      GE2 & 2.74$^\circ$ & 5.19$^\circ$ & 3.26$^\circ$ & 1.11$^\circ$ &13.51$^\circ$ \\
      DBPCA & 2.83$^\circ$ & 6.41$^\circ$ & 3.69$^\circ$ & 0.47$^\circ$ & 18.34$^\circ$\\
      DOCC(sum) & 4.93$^\circ$& 6.25$^\circ$& 5.50$^\circ$& 1.76$^\circ$ &12.48$^\circ$\\
DOCC(max) & 2.40$^\circ$& 5.46$^\circ$& 3.33$^\circ$& 0.41$^\circ$ &15.34$^\circ$\\
     GM($n$jet) & 2.28$^\circ$ & 3.92$^\circ$ & 2.70$^\circ$ & 0.52$^\circ$& 9.91$^\circ$ \\
   GM(pixel) & 2.27$^\circ$ & 3.70$^\circ$ & 2.53$^\circ$ & 0.46$^\circ$ &9.32$^\circ$ \\
SVR & 2.17$^\circ$ & \textbf{--} & \textbf{--} & \textbf{--} & \textbf{--}\\
CM(3 edge) & 3.60$^\circ$ & 4.10$^\circ$ & \textbf{--} & \textbf{--} & \textbf{--}\\
CM(9 edge) & 2.00$^\circ$ & 2.60$^\circ$ & \textbf{--} & \textbf{--} & \textbf{--}\\
   \hline
        IIC & 8.23$^\circ$ & $15.53^\circ$ & $10.72^\circ$ & 2.24$^\circ$ & 40.23$^\circ$ \\
NICs & 2.68$^\circ$ & \textbf{--} & \textbf{--} & \textbf{--} & \textbf{--}\\
WGE & 2.44$^\circ$ & 5.59$^\circ$ & 2.90$^\circ$ & 0.70$^\circ$ &16.00$^\circ$  \\
Zeta-image & 1.90$^\circ$ & 4.30$^\circ$ & \textbf{--} & \textbf{--}& \textbf{--} \\
    \hline
      DCs & 1.71$^\circ$ & 4.21$^\circ$ & 2.45$^\circ$ & 0.41$^\circ$ & 12.12$^\circ$ \\
    \hline
  \end{tabular}
\end{table*}

\begin{table*}[!t]
  \centering
  \small
  \captionsetup{justification=centering}
  \caption{WST test on the SFU laboratory database.}\label{WST_sfulab}
  \begin{tabular}{ !{\vrule width1.2pt}l!{\vrule width1.2pt}c!{\vrule width1.2pt}c
  !{\vrule width1.2pt}c!{\vrule width1.2pt}c!{\vrule width1.2pt}c
  !{\vrule width1.2pt}c!{\vrule width1.2pt}c!{\vrule width1.2pt}c
  !{\vrule width1.2pt}c!{\vrule width1.2pt}c!{\vrule width1.2pt}c
  !{\vrule width1.2pt}c!{\vrule width1.2pt}c!{\vrule width1.2pt}c
  !{\vrule width1.2pt}c!{\vrule width1.2pt}c!{\vrule width1.2pt}c!{\vrule width1.2pt}}

  \Xhline{1.2pt}
   & \rotatebox{90}{(1)DN}  & \rotatebox{90}{(2)IIC} & \rotatebox{90}{(3)GW} & \rotatebox{90}{(4)WP}
   & \rotatebox{90}{(5)DOCC(sum)} & \rotatebox{90}{(6)SS} & \rotatebox{90}{(7)SoG} & \rotatebox{90}{(8)GE1}
   & \rotatebox{90}{(9)GE2} & \rotatebox{90}{(10)GG} & \rotatebox{90}{(11)NUS} & \rotatebox{90}{(12)WGE}
   & \rotatebox{90}{(13)DOCC(max)} & \rotatebox{90}{(14)GM($n$jet)} & \rotatebox{90}{((15)GM(pixel)} & \rotatebox{90}{((15)DCs} & \rotatebox{90}{Score}\\
   \Xhline{1.2pt}
  \cellcolor{gray!25}{(1)}
  & \cellcolor{yellow!25}{0} &    \cellcolor{red!25}{-1}   &     \cellcolor{red!25}{-1}
  &     \cellcolor{red!25}{-1}   &     \cellcolor{red!25}{-1}   &     \cellcolor{red!25}{-1}
  &     \cellcolor{red!25}{-1}   &     \cellcolor{red!25}{-1}   &     \cellcolor{red!25}{-1}
  &     \cellcolor{red!25}{-1}   &     \cellcolor{red!25}{-1}   &     \cellcolor{red!25}{-1}
  &     \cellcolor{red!25}{-1}   &     \cellcolor{red!25}{-1}   &     \cellcolor{red!25}{-1}
  &     \cellcolor{red!25}{-1}   &  \cellcolor{gray!25}{0}\\
  \Xhline{1.2pt}
  \cellcolor{gray!25}{(2)}
  & \cellcolor{green!25}{1} &      \cellcolor{yellow!25}{0} &    \cellcolor{red!25}{-1}
  &     \cellcolor{red!25}{-1}   &     \cellcolor{red!25}{-1}   &     \cellcolor{red!25}{-1}
  &     \cellcolor{red!25}{-1}   &     \cellcolor{red!25}{-1}   &     \cellcolor{red!25}{-1}
  &     \cellcolor{red!25}{-1}   &     \cellcolor{red!25}{-1}   &     \cellcolor{red!25}{-1}
  &     \cellcolor{red!25}{-1}   &     \cellcolor{red!25}{-1}   &     \cellcolor{red!25}{-1}
  &     \cellcolor{red!25}{-1}   &  \cellcolor{gray!25}{1}\\
  \Xhline{1.2pt}
  \cellcolor{gray!25}{(3)}
  & \cellcolor{green!25}{1} &      \cellcolor{green!25}{1} &      \cellcolor{yellow!25}{0}
  &     \cellcolor{yellow!25}{0} &    \cellcolor{red!25}{-1}   &     \cellcolor{red!25}{-1}
  &     \cellcolor{red!25}{-1}   &     \cellcolor{red!25}{-1}   &     \cellcolor{red!25}{-1}
  &     \cellcolor{red!25}{-1}   &     \cellcolor{red!25}{-1}   &     \cellcolor{red!25}{-1}
  &     \cellcolor{red!25}{-1}   &     \cellcolor{red!25}{-1}   &     \cellcolor{red!25}{-1}
  &     \cellcolor{red!25}{-1}   & \cellcolor{gray!25}{2}\\
  \Xhline{1.2pt}
 \cellcolor{gray!25}{(4)}
  & \cellcolor{green!25}{1} &      \cellcolor{green!25}{1} &      \cellcolor{yellow!25}{0}
  &     \cellcolor{yellow!25}{0} &    \cellcolor{red!25}{-1}   &     \cellcolor{red!25}{-1}
  &     \cellcolor{red!25}{-1}   &     \cellcolor{red!25}{-1}   &     \cellcolor{red!25}{-1}
  &     \cellcolor{red!25}{-1}   &     \cellcolor{red!25}{-1}   &     \cellcolor{red!25}{-1}
  &     \cellcolor{red!25}{-1}   &     \cellcolor{red!25}{-1}   &     \cellcolor{red!25}{-1}
  &     \cellcolor{red!25}{-1}   &  \cellcolor{gray!25}{2}\\
  \Xhline{1.2pt}
\cellcolor{gray!25}{(5)}
  &  \cellcolor{green!25}{1} &      \cellcolor{green!25}{1} &      \cellcolor{green!25}{1}
  &      \cellcolor{green!25}{1} &      \cellcolor{yellow!25}{0} &    \cellcolor{red!25}{-1}
  &     \cellcolor{red!25}{-1}   &     \cellcolor{red!25}{-1}   &     \cellcolor{red!25}{-1}
  &     \cellcolor{red!25}{-1}   &     \cellcolor{red!25}{-1}   &     \cellcolor{red!25}{-1}
  &     \cellcolor{red!25}{-1}   &     \cellcolor{red!25}{-1}   &     \cellcolor{red!25}{-1}
  &     \cellcolor{red!25}{-1}   & \cellcolor{gray!25}{4}\\
  \Xhline{1.2pt}
  \cellcolor{gray!25}{(6)}
  & \cellcolor{green!25}{1} &      \cellcolor{green!25}{1} &      \cellcolor{green!25}{1}
  &      \cellcolor{green!25}{1} &      \cellcolor{green!25}{1} &      \cellcolor{yellow!25}{0}
  &     \cellcolor{yellow!25}{0} &     \cellcolor{yellow!25}{0} &    \cellcolor{red!25}{-1}
  &     \cellcolor{red!25}{-1}   &     \cellcolor{red!25}{-1}   &     \cellcolor{red!25}{-1}
  &     \cellcolor{red!25}{-1}   &     \cellcolor{red!25}{-1}   &     \cellcolor{red!25}{-1}
  &     \cellcolor{red!25}{-1}   &  \cellcolor{gray!25}{5}\\
  \Xhline{1.2pt}
  \cellcolor{gray!25}{(7)}
 &  \cellcolor{green!25}{1} &      \cellcolor{green!25}{1} &      \cellcolor{green!25}{1}
 &      \cellcolor{green!25}{1} &      \cellcolor{green!25}{1} &      \cellcolor{yellow!25}{0}
 &     \cellcolor{yellow!25}{0} &     \cellcolor{yellow!25}{0} &     \cellcolor{yellow!25}{0}
 &     \cellcolor{yellow!25}{0} &     \cellcolor{yellow!25}{0} &    \cellcolor{red!25}{-1}
 &     \cellcolor{red!25}{-1}   &     \cellcolor{red!25}{-1}   &     \cellcolor{red!25}{-1}
 &     \cellcolor{red!25}{-1}   &  \cellcolor{gray!25}{5}\\
 \Xhline{1.2pt}
  \cellcolor{gray!25}{(8)}
  & \cellcolor{green!25}{1} &      \cellcolor{green!25}{1} &      \cellcolor{green!25}{1}
  &      \cellcolor{green!25}{1} &      \cellcolor{green!25}{1} &      \cellcolor{yellow!25}{0}
  &     \cellcolor{yellow!25}{0} &     \cellcolor{yellow!25}{0} &     \cellcolor{yellow!25}{0}
  &     \cellcolor{yellow!25}{0} &     \cellcolor{yellow!25}{0} &    \cellcolor{red!25}{-1}
  &     \cellcolor{red!25}{-1}   &     \cellcolor{red!25}{-1}   &     \cellcolor{red!25}{-1}
  &     \cellcolor{red!25}{-1}   &  \cellcolor{gray!25}{5}\\
    \Xhline{1.2pt}
  \cellcolor{gray!25}{(9)}
  & \cellcolor{green!25}{1} &      \cellcolor{green!25}{1} &      \cellcolor{green!25}{1}
  &      \cellcolor{green!25}{1} &      \cellcolor{green!25}{1} &      \cellcolor{green!25}{1}
  &      \cellcolor{yellow!25}{0} &     \cellcolor{yellow!25}{0} &     \cellcolor{yellow!25}{0}
  &     \cellcolor{yellow!25}{0} &     \cellcolor{yellow!25}{0} &    \cellcolor{red!25}{-1}
  &     \cellcolor{red!25}{-1}   &     \cellcolor{red!25}{-1}   &     \cellcolor{red!25}{-1}
  &     \cellcolor{red!25}{-1}   & \cellcolor{gray!25}{6}\\
    \Xhline{1.2pt}
  \cellcolor{gray!25}{(10)}
  & \cellcolor{green!25}{1} &      \cellcolor{green!25}{1} &      \cellcolor{green!25}{1}
  &      \cellcolor{green!25}{1} &      \cellcolor{green!25}{1} &      \cellcolor{green!25}{1}
  &      \cellcolor{yellow!25}{0} &     \cellcolor{yellow!25}{0} &     \cellcolor{yellow!25}{0}
  &     \cellcolor{yellow!25}{0} &     \cellcolor{yellow!25}{0} &     \cellcolor{yellow!25}{0}
  &     \cellcolor{yellow!25}{0} &    \cellcolor{red!25}{-1}   &     \cellcolor{red!25}{-1}
  &     \cellcolor{red!25}{-1}   &  \cellcolor{gray!25}{6}\\
    \Xhline{1.2pt}
   \cellcolor{gray!25}{(11)}
  & \cellcolor{green!25}{1} &      \cellcolor{green!25}{1} &      \cellcolor{green!25}{1}
  &      \cellcolor{green!25}{1} &      \cellcolor{green!25}{1} &      \cellcolor{green!25}{1}
  &      \cellcolor{yellow!25}{0} &     \cellcolor{yellow!25}{0} &     \cellcolor{yellow!25}{0}
  &     \cellcolor{yellow!25}{0} &     \cellcolor{yellow!25}{0} &     \cellcolor{yellow!25}{0}
  &     \cellcolor{yellow!25}{0} &    \cellcolor{red!25}{-1}   &     \cellcolor{red!25}{-1}
  &     \cellcolor{red!25}{-1}   &  \cellcolor{gray!25}{6}\\
    \Xhline{1.2pt}
  \cellcolor{gray!25}{(12)}
  & \cellcolor{green!25}{1} &      \cellcolor{green!25}{1} &      \cellcolor{green!25}{1}
  &      \cellcolor{green!25}{1} &      \cellcolor{green!25}{1} &      \cellcolor{green!25}{1}
  &      \cellcolor{green!25}{1} &      \cellcolor{green!25}{1} &      \cellcolor{green!25}{1}
  &      \cellcolor{yellow!25}{0} &     \cellcolor{yellow!25}{0} &     \cellcolor{yellow!25}{0}
  &     \cellcolor{yellow!25}{0} &     \cellcolor{yellow!25}{0} &    \cellcolor{red!25}{-1}
  &     \cellcolor{red!25}{-1}   &  \cellcolor{gray!25}{9}\\
    \Xhline{1.2pt}
  \cellcolor{gray!25}{(13)}
  &  \cellcolor{green!25}{1} &      \cellcolor{green!25}{1} &      \cellcolor{green!25}{1}
   &      \cellcolor{green!25}{1} &      \cellcolor{green!25}{1} &      \cellcolor{green!25}{1}
    &      \cellcolor{green!25}{1} &      \cellcolor{green!25}{1} &      \cellcolor{green!25}{1}
    &      \cellcolor{yellow!25}{0} &     \cellcolor{yellow!25}{0} &     \cellcolor{yellow!25}{0}
    &     \cellcolor{yellow!25}{0} &     \cellcolor{yellow!25}{0} &     \cellcolor{yellow!25}{0}
    &    \cellcolor{yellow!25}{0}   &  \cellcolor{gray!25}{9}\\
    \Xhline{1.2pt}
  \cellcolor{gray!25}{(14)}
  & \cellcolor{green!25}{1} &      \cellcolor{green!25}{1} &      \cellcolor{green!25}{1}
  &      \cellcolor{green!25}{1} &      \cellcolor{green!25}{1} &      \cellcolor{green!25}{1}
  &      \cellcolor{green!25}{1} &      \cellcolor{green!25}{1} &      \cellcolor{green!25}{1}
  &      \cellcolor{green!25}{1} &      \cellcolor{green!25}{1} &      \cellcolor{yellow!25}{0}
  &     \cellcolor{yellow!25}{0} &     \cellcolor{yellow!25}{0} &     \cellcolor{yellow!25}{0}
  &    \cellcolor{yellow!25}{0}   &  \cellcolor{gray!25}{11}\\
    \Xhline{1.2pt}
  \cellcolor{gray!25}{(15)}
  & \cellcolor{green!25}{1} &      \cellcolor{green!25}{1} &      \cellcolor{green!25}{1}
  &      \cellcolor{green!25}{1} &      \cellcolor{green!25}{1} &      \cellcolor{green!25}{1}
  &      \cellcolor{green!25}{1} &      \cellcolor{green!25}{1} &      \cellcolor{green!25}{1}
  &      \cellcolor{green!25}{1} &      \cellcolor{green!25}{1} &      \cellcolor{green!25}{1}
  &      \cellcolor{yellow!25}{0} &     \cellcolor{yellow!25}{0} &     \cellcolor{yellow!25}{0}
  &     \cellcolor{yellow!25}{0} &\cellcolor{gray!25}{12}\\
  \Xhline{1.2pt}
  \cellcolor{gray!25}{(16)}
  & \cellcolor{green!25}{1} &      \cellcolor{green!25}{1} &      \cellcolor{green!25}{1}
  &      \cellcolor{green!25}{1} &      \cellcolor{green!25}{1} &      \cellcolor{green!25}{1}
  &      \cellcolor{green!25}{1} &      \cellcolor{green!25}{1} &      \cellcolor{green!25}{1}
  &      \cellcolor{green!25}{1} &      \cellcolor{green!25}{1} &      \cellcolor{green!25}{1}
  &      \cellcolor{yellow!25}{0} &      \cellcolor{yellow!25}{0} &      \cellcolor{yellow!25}{0}
  &     \cellcolor{yellow!25}{0} & \cellcolor{gray!25}{12}\\
  \Xhline{1.2pt}
\end{tabular}
\end{table*}

Throughout the experimental evaluations, we extract the derivative colors with scales $\sigma=1,2$ for the gaussian operators.
And the kernel bandwidth is set to be $h=0.03$. In addition, to simply the algorithm, we detect the
original regions with the top $5\%$ bright pixels for all databases, i.e., setting $\tau$ to be 5.
To this end, the only undetermined parameter in the proposed approach is
$\eta$, which is used to set the intensity threshold.
We demonstrate the stable performance of DCs under different settings of $\eta$ on each database with experiments.
The average runtime per-image is reported on each database to show the efficiency
of the proposed approach. We will later
release our code with the paper.
\subsection{SFU Laboratory Database}
\begin{figure*}[!t]
  \centering
  \subfigure[] {\includegraphics[width=50mm]{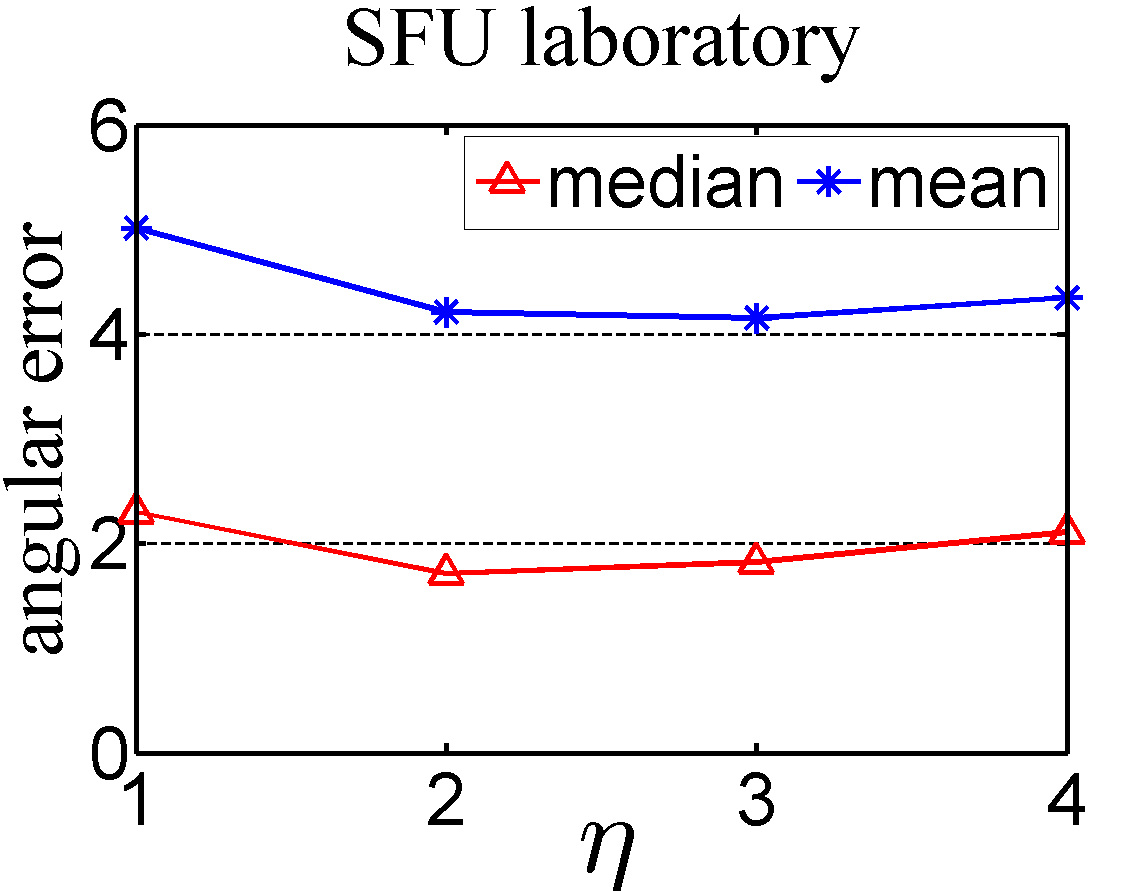}\label{sfu_vary}}
  \subfigure[] {\includegraphics[width=50mm]{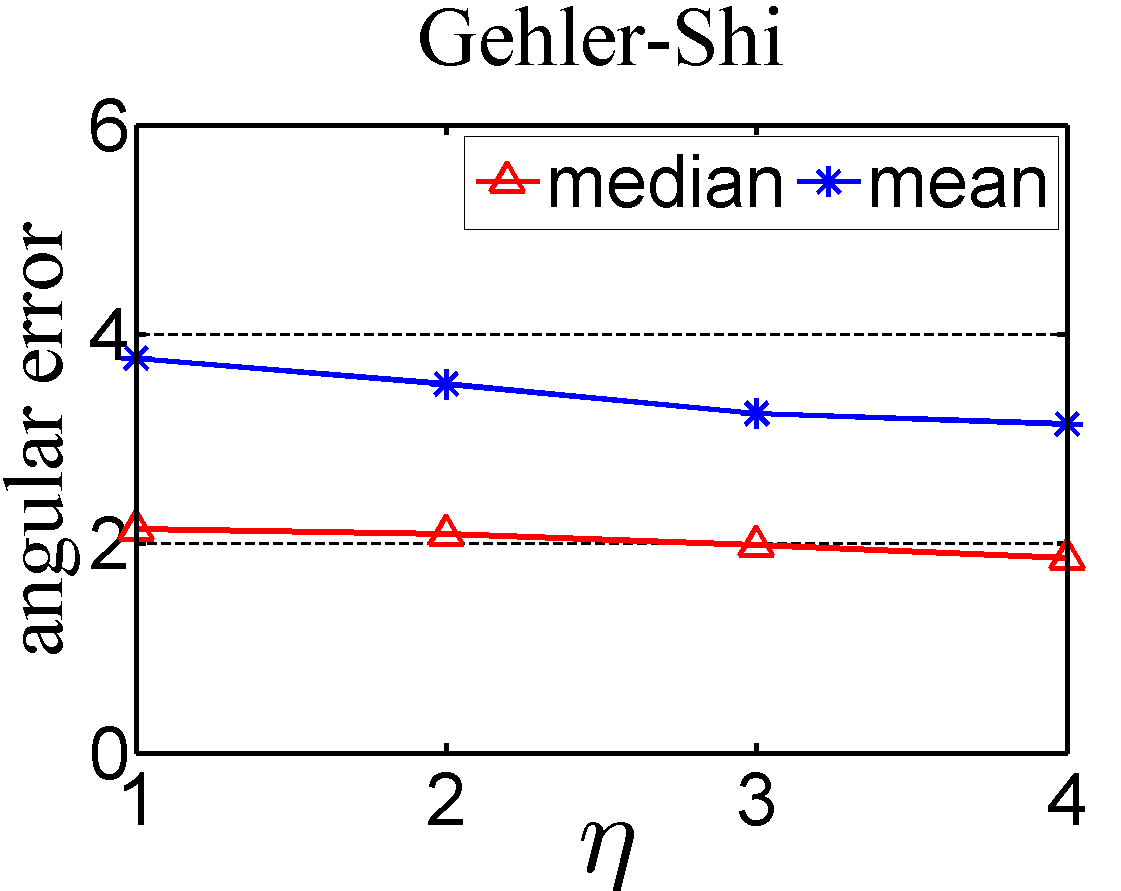}\label{shi_vary}}
  \subfigure[] {\includegraphics[width=50mm]{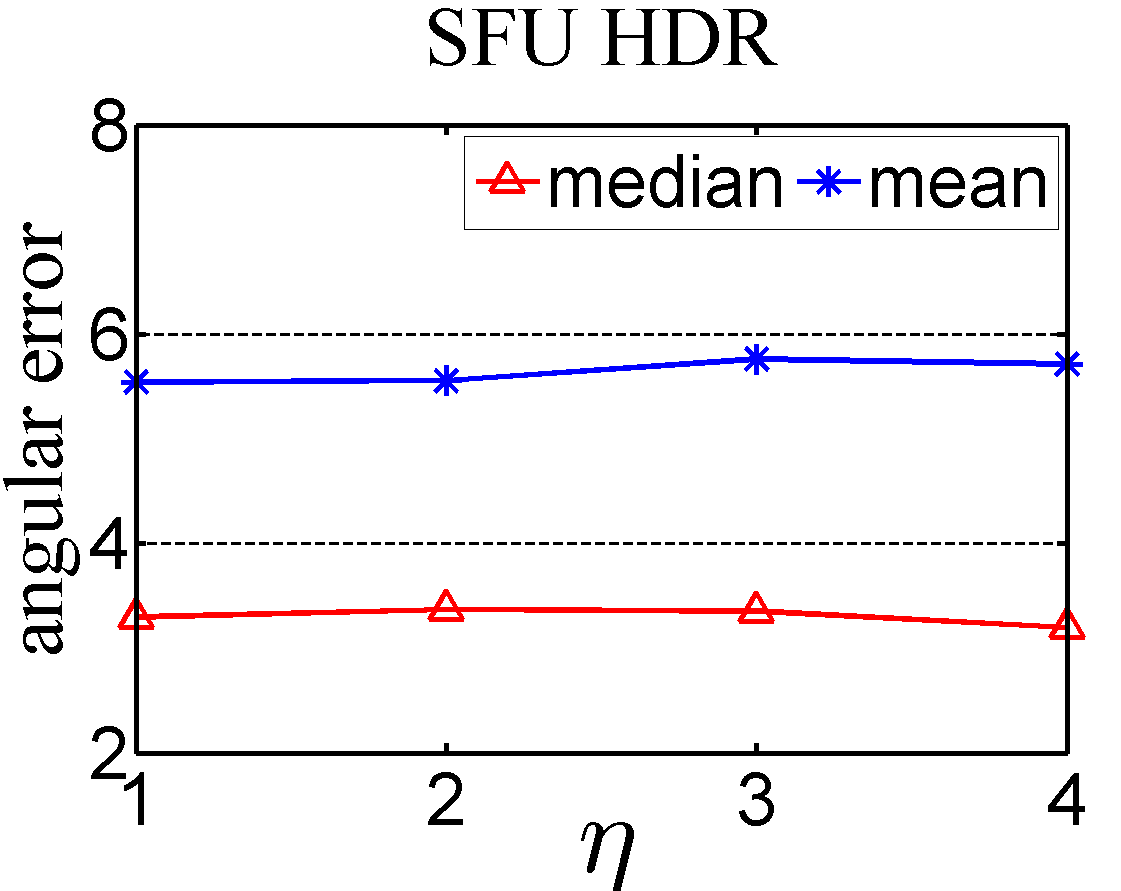}\label{hdr_vary}}
  \caption{The relationship between performance of the proposed approach on the different databases and the parameter $\eta$.
  It can be seen that the proposed approach performs quite stably while $\eta$ is ranging from 1 to 4.}
\end{figure*}


The SFU laboratory database contains totally 321 images of 31 different scenes, which are recorded
under 11 different illuminants in laboratory settings. Among these images, 223 are with minimal specularities,
while 98 are with non-negligible dielectric specularities, which consequently makes this database quite suitable for
evaluating the proposed approach.

We firstly track how the the proposed approach performs under variations of parameter $\eta$ with experiments.
Fig. \ref{sfu_vary} shows the results.
It can be noticed that performance of the proposed approach keeps quite stable while $\eta$ is ranging from 1 to 4.
The results reported in Table \ref{sfulab} and Table \ref{WST_sfulab} are obtained with $\eta=2$.
To demonstrate that the erosion operation is necessary, we conduct similar experiment
by directly selecting $\mathcal{L}^h$ as locations of the top 5\% bright pixels.
Compared with the results shows in Fig. \ref{sfu_vary}, the median and mean angular errors degrade to be $3.28^\circ$ and 5.07$^\circ$ respectively.
Thus,
removing the small regions in the binary mask with an erosion operation improves the estimation accuracy.

Statistical metrics of both the proposed approach and the other ones (under their optimal parameter settings) are reported in
Table \ref{sfulab}.
Table \ref{WST_sfulab} shows the results of the sign test, in which
we omit the analysis of SVR, NICs, CM and Zeta-image because their error distributions are unavailable. Particularly,
a sign (0) at location $(i,j)$
indicates that there is no significant difference between the median errors of method $i$ and method $j$ at the 95\% confidence level.
A sign (1) indicates that the median of method $i$ is significantly lower than method $j$, and a sign (-1) indicates the opposite situation. In addition, the score in the last column computes the times that a method performs significantly better than others.

To the best of our knowledge, the proposed approach provides the lowest median angular error (1.71$^\circ$) on the SFU laboratory database.
From Table \ref{sfulab}, it can seen that its trimean and the \emph{Best 25 percent} are also lower compared to those of state-of-the-art methods.
Table \ref{WST_sfulab} shows that DCs significantly outperforms most methods. In addition, it has no significant difference with
DOCC(max), GM($n$jet) and GM(pixel).
Besides satisfactory performance, the proposed approach has such important advantages as
no training phase and simple implementation.
In addition, with simple Matlab implementation on a laptop with the CPU Interl Core i5-4200U, our average runtime on the SFU laboratory database is
0.82 second for each image. This runtime is shorter than most methods with code provided. However, since the performance of
an approach is more important than the runtime, we only report the time here for a referee.

\subsection{Gehler-Shi Database}

We then apply the proposed approach to a database free of laboratory controls: the Gehler-Shi database \cite{shi2010data},
which is the reprocessed version of the Color Checker database created by the authors of \cite{gehler2008bayesian}.
This database includes totally 568 images, among which
246 are labeled as indoor scenes and 322 outdoor scenes.
More importantly, images of this database are all 12-bit
linear images, which are generated directly from their RAW formats
and therefore free of any color correction.
Each image contains a color checker with known coordinates in the image space,
which is masked during the illuminant estimation.
The black-level offset for camera Canon 1D is zero, and that for Canon 5D is 128.
Therefore, before estimating the illuminant color from images taken by Canon 5D,
the black offset 128 should be subtracted.
Similarly,
it can be noticed from Fig. \ref{shi_vary} that performance of the proposed approach keeps more stably while $\eta$
is ranging from 1 to 4. The results reported in Table \ref{shi} and Table \ref{WST_shi}
are obtained with $\eta=4$.

Table \ref{shi} reports performance of various methods on the indoor, outdoor and the entire Gehler-Shi databases.
The optimal results (with $n=6$) of DBPCA is reported by executing the provided by Cheng et al.
The results of WGE is obtained with the parameter setting $(\kappa,p,\sigma)=(50,1,2)$. In addition, to obtain
the results of GM($n$jet), we set the parameters $(n,\sigma)$ to be $(1,5)$,  and meanwhile adopt the training and test configurations
recommended in \cite{li2015multi, li2014evaluating}, which is also used to generate results of other algorithms that demand a training session. Moreover, the optimal parameter settings for DOCC(max) and DOCC(sum) are $(k,\sigma)=(0.1,8.0)$, $(k,\sigma)=(1.0,5.5)$ respectively.
We notice that DCs provides the lowest median and mean errors compared to state-of-the-art methods.
Further, its median on the indoor subset is the lowest as well, slightly better than that of WGE. In addition, it can be seen
that similar to most approaches, DCs yields better results on the outdoor subset
than on the indoor subset.
The dominant reason attributed to this outcome is interreflection
in the indoor scenes, which as a result making the neutral illuminant assumption or NIR assumption in the dichromatic reflection model being violated.
Chakrabarti et al. suggested a similar explanation for this
phenomenon \cite{chakrabarti2012color}.
Table \ref{WST_shi} shows the results of the sign test.
It can be seen that DCs significantly outperforms all the other methods, with the only
exception of MC.  However, compared to the complexity of MC, DCs is much simpler and more efficient.
Our
average runtime for processing an image is 5.58 seconds on the Gehler-Shi database.

\begin{table*}[!t]
\captionsetup{justification=centering}
\caption{Performance on the Gehler-Shi database.}\label{shi}
  \centering
  \begin{tabular}{|l|c|c|c|c|c|c|c|c|c|}
    \hline
   \multirow{2}{*}{Method}  & \multicolumn{3}{c|}{All Images (568)} & \multicolumn{3}{c|}{Indoor (246)} & \multicolumn{3}{c|}{Outdoor (322)}\\
  \cline{2-10}
    & Median & Mean &  Worst-25\% & Median & Mean  & Worst-25\% & Median & Mean &  Worst-25\%  \\
    \hline
   \hline
   DN  & 4.80$^\circ$ & 9.26$^\circ$ & 24.03$^\circ$
   & 17.68$^\circ$ & 17.23$^\circ$ & 28.30$^\circ$
   & 2.75$^\circ$ & 3.17$^\circ$ & 6.00$^\circ$ \\
    \hline

WP   & 9.15$^\circ$ &  10.26$^\circ$ &  20.51$^\circ$
&  11.33$^\circ$ &  11.64$^\circ$ &  22.05$^\circ$
&  6.44$^\circ$ &  9.21$^\circ$ &  19.18$^\circ$  \\

Bayesian   & 5.14$^\circ$ &  6.74$^\circ$ &  15.03$^\circ$
&  6.54$^\circ$ &  7.92$^\circ$ &  16.27$^\circ$
&  4.23$^\circ$ &  5.84$^\circ$ &  13.52$^\circ$ \\

SoG    & 4.48$^\circ$ &  6.42$^\circ$ &  15.01$^\circ$
&  6.88$^\circ$ &  7.74$^\circ$ &  16.26$^\circ$
&  3.07$^\circ$ &  5.41$^\circ$ &  13.58$^\circ$  \\

GM(pixel)     & 3.98$^\circ$ &  6.00$^\circ$ &  14.31$^\circ$
&  6.12$^\circ$ &  7.46$^\circ$ &  15.84$^\circ$
&  3.00$^\circ$ &  4.90$^\circ$ &  12.37$^\circ$ \\

GG     & 3.90$^\circ$ &  6.35$^\circ$ &  15.83$^\circ$
&  5.94$^\circ$ &  8.08$^\circ$ &  17.51$^\circ$
&  2.69$^\circ$ &  5.02$^\circ$ &  13.34$^\circ$  \\

NN & 3.77$^\circ$ &  5.16$^\circ$ &  11.53$^\circ$
&  5.72$^\circ$ &  7.41$^\circ$ &  15.25$^\circ$
&  2.81$^\circ$ &  3.45$^\circ$ &  7.11$^\circ$ \\

GM($1$jet)  & 3.68$^\circ$ &  5.52$^\circ$ &  13.38$^\circ$
&  5.09$^\circ$ &  6.76$^\circ$ &  14.69$^\circ$
&  2.72$^\circ$ &  4.57$^\circ$ &  11.68$^\circ$  \\

GW   & 3.63$^\circ$ &  4.77$^\circ$ &  10.51$^\circ$
&  3.25$^\circ$ &  4.10$^\circ$ &  8.96$^\circ$
&  3.97$^\circ$ &  5.28$^\circ$ &  11.47$^\circ$  \\

GE1  & 3.28$^\circ$ &  4.19$^\circ$ &  8.75$^\circ$
&  2.97$^\circ$ &  3.73$^\circ$ &  8.09$^\circ$
&  3.58$^\circ$ &  4.54$^\circ$ &  9.15$^\circ$  \\

GE2  & 3.35$^\circ$ &  4.23$^\circ$ &  8.61$^\circ$
&  3.07$^\circ$ &  3.78$^\circ$ &  7.92$^\circ$
&  3.69$^\circ$ &  4.58$^\circ$ &  9.02$^\circ$  \\

SS  & 3.24$^\circ$ &  3.99$^\circ$ &  7.62$^\circ$
&  3.49$^\circ$ &  4.23$^\circ$ &  8.12$^\circ$
&  2.97$^\circ$ &  3.81$^\circ$ &  7.20$^\circ$  \\

SVR   & 3.23$^\circ$ &  4.14$^\circ$ &  9.05$^\circ$
& 4.46$^\circ$ &  5.66$^\circ$ &  11.91$^\circ$
& 2.47$^\circ$ &  2.97$^\circ$ &  5.99$^\circ$  \\

NIS  & 3.12$^\circ$ &  4.32$^\circ$ &  9.88$^\circ$
&  3.71$^\circ$ &  4.93$^\circ$ &  10.90$^\circ$
&  2.76$^\circ$ &  3.86$^\circ$ &  8.79$^\circ$  \\

HVI & 3.06$^\circ$ & 4.36$^\circ$ & 9.89$^\circ$
& 3.27$^\circ$ & 4.14$^\circ$ & 8.85$^\circ$
& 2.96$^\circ$ & 4.54$^\circ$ & 10.63$^\circ$  \\

DOCC(sum)   & 2.70$^\circ$ & 3.88$^\circ$ & 8.93$^\circ$
            & 2.84$^\circ$ & 3.78$^\circ$ & 8.28$^\circ$
            & 2.65$^\circ$ & 3.96$^\circ$ & 9.36$^\circ$ \\

DOCC(max)   & 2.69$^\circ$ & 4.64$^\circ$ & 11.85$^\circ$
& 4.22$^\circ$ & 5.99$^\circ$ & 14.23$^\circ$
& 2.02$^\circ$ & 3.61$^\circ$ & 9.23$^\circ$\\

NUS & 2.46$^\circ$ & 4.07$^\circ$ & 10.06$^\circ$
& 3.06$^\circ$ & 4.50$^\circ$ & 10.32$^\circ$
& 1.99$^\circ$ & 3.74$^\circ$ & 9.73$^\circ$ \\

MC & 2.13$^\circ$ & 3.25$^\circ$ & 8.09$^\circ$
& 3.37$^\circ$ & 4.42$^\circ$ & 9.97$^\circ$
& 1.39$^\circ$ & 2.35$^\circ$ & 6.02$^\circ$\\

%
%
       \hline
IIC & 4.53$^\circ$ & 8.07$^\circ$ & 20.88$^\circ$
  & 7.33$^\circ$ & 11.91$^\circ$ & 28.68$^\circ$
  & 3.49$^\circ$ & 5.13$^\circ$ & 11.85$^\circ$ \\

WGE & 2.51$^\circ$ & 3.50$^\circ$ & 8.00$^\circ$
& 2.10$^\circ$ & 3.17$^\circ$ & 7.73$^\circ$
& 2.91$^\circ$ & 3.76$^\circ$ & 8.01$^\circ$  \\

\hline
DCs & 1.86$^\circ$ &  3.14$^\circ$ &  7.93$^\circ$
&  2.06$^\circ$ &  3.53$^\circ$ &  8.99$^\circ$
&  1.68$^\circ$ &  2.84$^\circ$ &  7.04$^\circ$ \\
%

    \hline
  \end{tabular}
\end{table*}

\begin{table*}[!t]
  \centering
  \small
  \caption{WST test on the Gehler-Shi database.}\label{WST_shi}
  \begin{tabular}{!{\vrule width1.2pt}l!{\vrule width1.2pt}c!{\vrule width1.2pt}c
                  !{\vrule width1.2pt}c!{\vrule width1.2pt}c!{\vrule width1.2pt}c
                  !{\vrule width1.2pt}c!{\vrule width1.2pt}c!{\vrule width1.2pt}c
                  !{\vrule width1.2pt}c!{\vrule width1.2pt}c!{\vrule width1.2pt}c
                  !{\vrule width1.2pt}c!{\vrule width1.2pt}c!{\vrule width1.2pt}c
                  !{\vrule width1.2pt}c!{\vrule width1.2pt}c!{\vrule width1.2pt}c
                  !{\vrule width1.2pt}c!{\vrule width1.2pt}c!{\vrule width1.2pt}c
                  !{\vrule width1.2pt}c!{\vrule width1.2pt}c!{\vrule width1.2pt}c
                  !{\vrule width1.2pt}}

  \Xhline{1.2pt}
   & \rotatebox{90}{(1)WP}  & \rotatebox{90}{(2)DN} & \rotatebox{90}{(3)IIC} & \rotatebox{90}{(4)Bayesian}
   & \rotatebox{90}{(5)SoG} & \rotatebox{90}{(6)GM(pixel)} & \rotatebox{90}{(7)GG} & \rotatebox{90}{(8)NN}
   & \rotatebox{90}{(9)GM($1$jet)} & \rotatebox{90}{(10)GW} & \rotatebox{90}{(11)GE1} & \rotatebox{90}{(12)GE2}
   & \rotatebox{90}{(13)SS} & \rotatebox{90}{(14)HVI} & \rotatebox{90}{((15)SVR} &\rotatebox{90}{((16)NIS}
   & \rotatebox{90}{(17)DOCC(max)} & \rotatebox{90}{(18)DOCC(sum)} & \rotatebox{90}{(19)WGE}
   & \rotatebox{90}{(20)NUS} & \rotatebox{90}{(21)MC} & \rotatebox{90}{(22)DCs} & \rotatebox{90}{Score}\\
  \Xhline{1.2pt}
    \cellcolor{gray!25}{(1)}
    &   \cellcolor{yellow!25}{0} &    \cellcolor{red!25}{-1} &     \cellcolor{red!25}{-1}
    &     \cellcolor{red!25}{-1} &     \cellcolor{red!25}{-1} &     \cellcolor{red!25}{-1}
    &     \cellcolor{red!25}{-1} &     \cellcolor{red!25}{-1} &     \cellcolor{red!25}{-1}
    &     \cellcolor{red!25}{-1} &     \cellcolor{red!25}{-1} &     \cellcolor{red!25}{-1}
    &     \cellcolor{red!25}{-1} &     \cellcolor{red!25}{-1} &     \cellcolor{red!25}{-1}
    &     \cellcolor{red!25}{-1} &     \cellcolor{red!25}{-1} &     \cellcolor{red!25}{-1}
    &     \cellcolor{red!25}{-1} &     \cellcolor{red!25}{-1} &     \cellcolor{red!25}{-1}
    &     \cellcolor{red!25}{-1} &     \cellcolor{gray!25}{0}\\
    \Xhline{1.2pt}
  \cellcolor{gray!25}{(2)}
  &   \cellcolor{green!25}{1} &     \cellcolor{yellow!25}{0} &     \cellcolor{yellow!25}{0}
  &    \cellcolor{red!25}{-1} &     \cellcolor{red!25}{-1} &     \cellcolor{red!25}{-1}
  &     \cellcolor{red!25}{-1} &     \cellcolor{red!25}{-1} &     \cellcolor{red!25}{-1}
  &     \cellcolor{red!25}{-1} &     \cellcolor{red!25}{-1} &     \cellcolor{red!25}{-1}
  &     \cellcolor{red!25}{-1} &     \cellcolor{red!25}{-1} &     \cellcolor{red!25}{-1}
  &     \cellcolor{red!25}{-1} &     \cellcolor{red!25}{-1} &     \cellcolor{red!25}{-1}
  &     \cellcolor{red!25}{-1} &     \cellcolor{red!25}{-1} &     \cellcolor{red!25}{-1}
  &     \cellcolor{red!25}{-1} &    \cellcolor{gray!25}{1}\\
    \Xhline{1.2pt}
    \cellcolor{gray!25}{(3)}
  &   \cellcolor{green!25}{1} &     \cellcolor{yellow!25}{0} &     \cellcolor{yellow!25}{0}
  &     \cellcolor{yellow!25}{0} &    \cellcolor{red!25}{-1} &     \cellcolor{red!25}{-1}
  &     \cellcolor{red!25}{-1} &     \cellcolor{red!25}{-1} &     \cellcolor{red!25}{-1}
  &     \cellcolor{red!25}{-1} &     \cellcolor{red!25}{-1} &     \cellcolor{red!25}{-1}
  &     \cellcolor{red!25}{-1} &     \cellcolor{red!25}{-1} &     \cellcolor{red!25}{-1}
  &     \cellcolor{red!25}{-1} &     \cellcolor{red!25}{-1} &     \cellcolor{red!25}{-1}
  &     \cellcolor{red!25}{-1} &     \cellcolor{red!25}{-1} &     \cellcolor{red!25}{-1}
  &     \cellcolor{red!25}{-1} &    \cellcolor{gray!25}{1}\\
    \Xhline{1.2pt}
  \cellcolor{gray!25}{(4)}
  &   \cellcolor{green!25}{1} &     \cellcolor{green!25}{1} &     \cellcolor{yellow!25}{0}
  &     \cellcolor{yellow!25}{0} &     \cellcolor{yellow!25}{0} &    \cellcolor{red!25}{-1}
  &     \cellcolor{red!25}{-1} &     \cellcolor{red!25}{-1} &     \cellcolor{red!25}{-1}
  &     \cellcolor{red!25}{-1} &     \cellcolor{red!25}{-1} &     \cellcolor{red!25}{-1}
  &     \cellcolor{red!25}{-1} &     \cellcolor{red!25}{-1} &     \cellcolor{red!25}{-1}
  &     \cellcolor{red!25}{-1} &     \cellcolor{red!25}{-1} &     \cellcolor{red!25}{-1}
  &     \cellcolor{red!25}{-1} &     \cellcolor{red!25}{-1} &     \cellcolor{red!25}{-1}
  &     \cellcolor{red!25}{-1} &    \cellcolor{gray!25}{2}\\
    \Xhline{1.2pt}
  \cellcolor{gray!25}{(5)}
  &   \cellcolor{green!25}{1} &     \cellcolor{green!25}{1} &     \cellcolor{green!25}{1}
  &     \cellcolor{yellow!25}{0} &     \cellcolor{yellow!25}{0} &     \cellcolor{yellow!25}{0}
  &     \cellcolor{yellow!25}{0} &     \cellcolor{yellow!25}{0} &     \cellcolor{yellow!25}{0}
  &    \cellcolor{red!25}{-1} &     \cellcolor{red!25}{-1} &     \cellcolor{red!25}{-1}
  &     \cellcolor{red!25}{-1} &     \cellcolor{red!25}{-1} &     \cellcolor{red!25}{-1}
  &     \cellcolor{red!25}{-1} &     \cellcolor{red!25}{-1} &     \cellcolor{red!25}{-1}
  &     \cellcolor{red!25}{-1} &     \cellcolor{red!25}{-1} &     \cellcolor{red!25}{-1}
  &     \cellcolor{red!25}{-1} &     \cellcolor{gray!25}{3}\\
    \Xhline{1.2pt}
  \cellcolor{gray!25}{(6)}
  &   \cellcolor{green!25}{1} &     \cellcolor{green!25}{1} &     \cellcolor{green!25}{1}
  &     \cellcolor{green!25}{1} &     \cellcolor{yellow!25}{0} &     \cellcolor{yellow!25}{0}
  &     \cellcolor{yellow!25}{0} &     \cellcolor{yellow!25}{0} &     \cellcolor{yellow!25}{0}
  &    \cellcolor{red!25}{-1} &     \cellcolor{red!25}{-1} &     \cellcolor{red!25}{-1}
  &     \cellcolor{red!25}{-1} &     \cellcolor{red!25}{-1} &     \cellcolor{red!25}{-1}
  &     \cellcolor{red!25}{-1} &     \cellcolor{red!25}{-1} &     \cellcolor{red!25}{-1}
  &     \cellcolor{red!25}{-1} &     \cellcolor{red!25}{-1} &     \cellcolor{red!25}{-1}
  &     \cellcolor{red!25}{-1} &    \cellcolor{gray!25}{4}\\
    \Xhline{1.2pt}
  \cellcolor{gray!25}{(7)}
  &   \cellcolor{green!25}{1} &     \cellcolor{green!25}{1} &     \cellcolor{green!25}{1}
  &     \cellcolor{green!25}{1} &     \cellcolor{yellow!25}{0} &     \cellcolor{yellow!25}{0}
  &     \cellcolor{yellow!25}{0} &     \cellcolor{yellow!25}{0} &     \cellcolor{yellow!25}{0}
  &     \cellcolor{yellow!25}{0} &    \cellcolor{red!25}{-1} &     \cellcolor{red!25}{-1}
  &     \cellcolor{red!25}{-1} &     \cellcolor{red!25}{-1} &     \cellcolor{red!25}{-1}
  &     \cellcolor{red!25}{-1} &     \cellcolor{red!25}{-1} &     \cellcolor{red!25}{-1}
  &     \cellcolor{red!25}{-1} &     \cellcolor{red!25}{-1} &     \cellcolor{red!25}{-1}
  &     \cellcolor{red!25}{-1} &    \cellcolor{gray!25}{4}\\
    \Xhline{1.2pt}
  \cellcolor{gray!25}{(8)}
  &    \cellcolor{green!25}{1} &     \cellcolor{green!25}{1} &     \cellcolor{green!25}{1}
  &     \cellcolor{green!25}{1} &     \cellcolor{yellow!25}{0} &     \cellcolor{yellow!25}{0}
  &     \cellcolor{yellow!25}{0} &     \cellcolor{yellow!25}{0} &     \cellcolor{yellow!25}{0}
  &     \cellcolor{yellow!25}{0} &    \cellcolor{red!25}{-1} &     \cellcolor{red!25}{-1}
  &     \cellcolor{red!25}{-1} &     \cellcolor{red!25}{-1} &     \cellcolor{red!25}{-1}
  &     \cellcolor{red!25}{-1} &     \cellcolor{red!25}{-1} &     \cellcolor{red!25}{-1}
  &     \cellcolor{red!25}{-1} &     \cellcolor{red!25}{-1} &     \cellcolor{red!25}{-1}
  &     \cellcolor{red!25}{-1} &   \cellcolor{gray!25}{4}\\
    \Xhline{1.2pt}
  \cellcolor{gray!25}{(9)}
  &   \cellcolor{green!25}{1} &     \cellcolor{green!25}{1} &     \cellcolor{green!25}{1}
  &     \cellcolor{green!25}{1} &     \cellcolor{yellow!25}{0} &     \cellcolor{yellow!25}{0}
  &     \cellcolor{yellow!25}{0} &     \cellcolor{yellow!25}{0} &     \cellcolor{yellow!25}{0}
  &     \cellcolor{yellow!25}{0} &     \cellcolor{yellow!25}{0} &     \cellcolor{yellow!25}{0}
  &     \cellcolor{yellow!25}{0} &    \cellcolor{red!25}{-1} &     \cellcolor{red!25}{-1}
  &     \cellcolor{red!25}{-1} &     \cellcolor{red!25}{-1} &     \cellcolor{red!25}{-1}
  &     \cellcolor{red!25}{-1} &     \cellcolor{red!25}{-1} &     \cellcolor{red!25}{-1}
  &     \cellcolor{red!25}{-1} &    \cellcolor{gray!25}{4}\\
    \Xhline{1.2pt}
  \cellcolor{gray!25}{(10)}
  &   \cellcolor{green!25}{1} &     \cellcolor{green!25}{1} &     \cellcolor{green!25}{1}
  &     \cellcolor{green!25}{1} &     \cellcolor{green!25}{1} &     \cellcolor{green!25}{1}
  &     \cellcolor{yellow!25}{0} &     \cellcolor{yellow!25}{0} &     \cellcolor{yellow!25}{0}
  &     \cellcolor{yellow!25}{0} &     \cellcolor{yellow!25}{0} &     \cellcolor{yellow!25}{0}
  &     \cellcolor{yellow!25}{0} &     \cellcolor{yellow!25}{0} &    \cellcolor{red!25}{-1}
  &     \cellcolor{red!25}{-1} &     \cellcolor{red!25}{-1} &     \cellcolor{red!25}{-1}
  &     \cellcolor{red!25}{-1} &     \cellcolor{red!25}{-1} &     \cellcolor{red!25}{-1}
  &     \cellcolor{red!25}{-1} &     \cellcolor{gray!25}{6}\\
    \Xhline{1.2pt}
  \cellcolor{gray!25}{(11)}
   &   \cellcolor{green!25}{1} &     \cellcolor{green!25}{1} &     \cellcolor{green!25}{1}
   &     \cellcolor{green!25}{1} &     \cellcolor{green!25}{1} &     \cellcolor{green!25}{1}
   &     \cellcolor{green!25}{1} &     \cellcolor{green!25}{1} &     \cellcolor{yellow!25}{0}
   &     \cellcolor{yellow!25}{0} &     \cellcolor{yellow!25}{0} &     \cellcolor{yellow!25}{0}
   &     \cellcolor{yellow!25}{0} &     \cellcolor{yellow!25}{0} &     \cellcolor{yellow!25}{0}
   &     \cellcolor{yellow!25}{0} &    \cellcolor{red!25}{-1} &     \cellcolor{red!25}{-1}
   &     \cellcolor{red!25}{-1} &     \cellcolor{red!25}{-1} &     \cellcolor{red!25}{-1}
   &     \cellcolor{red!25}{-1} &     \cellcolor{gray!25}{8}\\
    \Xhline{1.2pt}
  \cellcolor{gray!25}{(12)}
  &   \cellcolor{green!25}{1} &     \cellcolor{green!25}{1} &     \cellcolor{green!25}{1}
  &     \cellcolor{green!25}{1} &     \cellcolor{green!25}{1} &     \cellcolor{green!25}{1}
  &     \cellcolor{green!25}{1} &     \cellcolor{green!25}{1} &     \cellcolor{yellow!25}{0}
  &     \cellcolor{yellow!25}{0} &     \cellcolor{yellow!25}{0} &     \cellcolor{yellow!25}{0}
  &     \cellcolor{yellow!25}{0} &     \cellcolor{yellow!25}{0} &     \cellcolor{yellow!25}{0}
  &     \cellcolor{yellow!25}{0} &    \cellcolor{red!25}{-1} &     \cellcolor{red!25}{-1}
  &     \cellcolor{red!25}{-1} &     \cellcolor{red!25}{-1} &     \cellcolor{red!25}{-1}
  &     \cellcolor{red!25}{-1} &     \cellcolor{gray!25}{8}\\
    \Xhline{1.2pt}
   \cellcolor{gray!25}{(13)}
   &   \cellcolor{green!25}{1} &     \cellcolor{green!25}{1} &     \cellcolor{green!25}{1}
   &     \cellcolor{green!25}{1} &     \cellcolor{green!25}{1} &     \cellcolor{green!25}{1}
   &     \cellcolor{green!25}{1} &     \cellcolor{green!25}{1} &     \cellcolor{yellow!25}{0}
   &     \cellcolor{yellow!25}{0} &     \cellcolor{yellow!25}{0} &     \cellcolor{yellow!25}{0}
   &     \cellcolor{yellow!25}{0} &     \cellcolor{yellow!25}{0} &     \cellcolor{yellow!25}{0}
   &     \cellcolor{yellow!25}{0} &    \cellcolor{red!25}{-1} &     \cellcolor{red!25}{-1}
   &     \cellcolor{red!25}{-1} &     \cellcolor{red!25}{-1} &     \cellcolor{red!25}{-1}
   &     \cellcolor{red!25}{-1} &    \cellcolor{gray!25}{8}\\
    \Xhline{1.2pt}
   \cellcolor{gray!25}{(14)}
   &   \cellcolor{green!25}{1} &     \cellcolor{green!25}{1} &     \cellcolor{green!25}{1}
   &     \cellcolor{green!25}{1} &     \cellcolor{green!25}{1} &     \cellcolor{green!25}{1}
   &     \cellcolor{green!25}{1} &     \cellcolor{green!25}{1} &     \cellcolor{green!25}{1}
   &     \cellcolor{yellow!25}{0} &     \cellcolor{yellow!25}{0} &     \cellcolor{yellow!25}{0}
   &     \cellcolor{yellow!25}{0} &     \cellcolor{yellow!25}{0} &     \cellcolor{yellow!25}{0}
   &     \cellcolor{yellow!25}{0} &    \cellcolor{red!25}{-1} &     \cellcolor{red!25}{-1}
   &     \cellcolor{red!25}{-1} &     \cellcolor{red!25}{-1} &     \cellcolor{red!25}{-1}
   &     \cellcolor{red!25}{-1} &    \cellcolor{gray!25}{9}\\
    \Xhline{1.2pt}
   \cellcolor{gray!25}{(15)}
   &   \cellcolor{green!25}{1} &     \cellcolor{green!25}{1} &     \cellcolor{green!25}{1}
   &     \cellcolor{green!25}{1} &     \cellcolor{green!25}{1} &     \cellcolor{green!25}{1}
   &     \cellcolor{green!25}{1} &     \cellcolor{green!25}{1} &     \cellcolor{green!25}{1}
   &     \cellcolor{green!25}{1} &     \cellcolor{yellow!25}{0} &     \cellcolor{yellow!25}{0}
    &     \cellcolor{yellow!25}{0} &     \cellcolor{yellow!25}{0} &     \cellcolor{yellow!25}{0}
     &     \cellcolor{yellow!25}{0} &     \cellcolor{yellow!25}{0} &    \cellcolor{red!25}{-1}
     &     \cellcolor{red!25}{-1} &     \cellcolor{red!25}{-1} &     \cellcolor{red!25}{-1}
     &     \cellcolor{red!25}{-1} &    \cellcolor{gray!25}{10}\\
    \Xhline{1.2pt}
   \cellcolor{gray!25}{(16)}
   &   \cellcolor{green!25}{1} &     \cellcolor{green!25}{1} &     \cellcolor{green!25}{1}
   &     \cellcolor{green!25}{1} &     \cellcolor{green!25}{1} &     \cellcolor{green!25}{1}
   &     \cellcolor{green!25}{1} &     \cellcolor{green!25}{1} &     \cellcolor{green!25}{1}
   &     \cellcolor{green!25}{1} &     \cellcolor{yellow!25}{0} &     \cellcolor{yellow!25}{0}
   &     \cellcolor{yellow!25}{0} &     \cellcolor{yellow!25}{0} &     \cellcolor{yellow!25}{0}
   &     \cellcolor{yellow!25}{0} &     \cellcolor{yellow!25}{0} &    \cellcolor{red!25}{-1}
   &     \cellcolor{red!25}{-1} &     \cellcolor{red!25}{-1} &     \cellcolor{red!25}{-1}
   &     \cellcolor{red!25}{-1} &   \cellcolor{gray!25}{10}\\
    \Xhline{1.2pt}
   \cellcolor{gray!25}{(17)}
   &   \cellcolor{green!25}{1} &     \cellcolor{green!25}{1} &     \cellcolor{green!25}{1}
   &     \cellcolor{green!25}{1} &     \cellcolor{green!25}{1} &     \cellcolor{green!25}{1}
   &     \cellcolor{green!25}{1} &     \cellcolor{green!25}{1} &     \cellcolor{green!25}{1}
   &     \cellcolor{green!25}{1} &     \cellcolor{green!25}{1} &     \cellcolor{green!25}{1}
   &     \cellcolor{green!25}{1} &     \cellcolor{green!25}{1} &     \cellcolor{yellow!25}{0}
   &     \cellcolor{yellow!25}{0} &     \cellcolor{yellow!25}{0} &     \cellcolor{yellow!25}{0}
   &     \cellcolor{yellow!25}{0} &     \cellcolor{yellow!25}{0} &    \cellcolor{red!25}{-1}
   &     \cellcolor{red!25}{-1} &     \cellcolor{gray!25}{14}\\
    \Xhline{1.2pt}
    \cellcolor{gray!25}{(18)}
    &   \cellcolor{green!25}{1} &     \cellcolor{green!25}{1} &     \cellcolor{green!25}{1}
    &     \cellcolor{green!25}{1} &     \cellcolor{green!25}{1} &     \cellcolor{green!25}{1}
    &     \cellcolor{green!25}{1} &     \cellcolor{green!25}{1} &     \cellcolor{green!25}{1}
    &     \cellcolor{green!25}{1} &     \cellcolor{green!25}{1} &     \cellcolor{green!25}{1}
    &     \cellcolor{green!25}{1} &     \cellcolor{green!25}{1} &     \cellcolor{green!25}{1}
    &     \cellcolor{green!25}{1} &     \cellcolor{yellow!25}{0} &     \cellcolor{yellow!25}{0}
    &     \cellcolor{yellow!25}{0} &     \cellcolor{yellow!25}{0} &    \cellcolor{red!25}{-1}
    &     \cellcolor{red!25}{-1} &   \cellcolor{gray!25}{16}\\
    \Xhline{1.2pt}
    \cellcolor{gray!25}{(19)}
    &   \cellcolor{green!25}{1} &     \cellcolor{green!25}{1} &     \cellcolor{green!25}{1}
    &     \cellcolor{green!25}{1} &     \cellcolor{green!25}{1} &     \cellcolor{green!25}{1}
    &     \cellcolor{green!25}{1} &     \cellcolor{green!25}{1} &     \cellcolor{green!25}{1}
    &     \cellcolor{green!25}{1} &     \cellcolor{green!25}{1} &     \cellcolor{green!25}{1}
    &     \cellcolor{green!25}{1} &     \cellcolor{green!25}{1} &     \cellcolor{green!25}{1}
    &     \cellcolor{green!25}{1} &     \cellcolor{yellow!25}{0} &     \cellcolor{yellow!25}{0}
    &     \cellcolor{yellow!25}{0} &     \cellcolor{yellow!25}{0} &    \cellcolor{red!25}{-1}
    &      \cellcolor{red!25}{-1} &    \cellcolor{gray!25}{16}\\
    \Xhline{1.2pt}
    \cellcolor{gray!25}{(20)}
    &   \cellcolor{green!25}{1} &     \cellcolor{green!25}{1} &     \cellcolor{green!25}{1}
    &     \cellcolor{green!25}{1} &     \cellcolor{green!25}{1} &     \cellcolor{green!25}{1}
    &     \cellcolor{green!25}{1} &     \cellcolor{green!25}{1} &     \cellcolor{green!25}{1}
    &     \cellcolor{green!25}{1} &     \cellcolor{green!25}{1} &     \cellcolor{green!25}{1}
    &     \cellcolor{green!25}{1} &     \cellcolor{green!25}{1} &     \cellcolor{green!25}{1}
    &     \cellcolor{green!25}{1} &     \cellcolor{yellow!25}{0} &     \cellcolor{yellow!25}{0}
    &     \cellcolor{yellow!25}{0} &     \cellcolor{yellow!25}{0} &    \cellcolor{red!25}{-1}
    &      \cellcolor{red!25}{-1} &    \cellcolor{gray!25}{16}\\
    \Xhline{1.2pt}
    \cellcolor{gray!25}{(21)}
    &   \cellcolor{green!25}{1} &     \cellcolor{green!25}{1} &     \cellcolor{green!25}{1}
    &     \cellcolor{green!25}{1} &     \cellcolor{green!25}{1} &     \cellcolor{green!25}{1}
    &     \cellcolor{green!25}{1} &     \cellcolor{green!25}{1} &     \cellcolor{green!25}{1}
    &     \cellcolor{green!25}{1} &     \cellcolor{green!25}{1} &     \cellcolor{green!25}{1}
    &     \cellcolor{green!25}{1} &     \cellcolor{green!25}{1} &     \cellcolor{green!25}{1}
    &     \cellcolor{green!25}{1} &     \cellcolor{green!25}{1} &     \cellcolor{green!25}{1}
    &     \cellcolor{green!25}{1} &     \cellcolor{green!25}{1} &     \cellcolor{yellow!25}{0}
    &     \cellcolor{yellow!25}{0} &    \cellcolor{gray!25}{20}\\
    \Xhline{1.2pt}
    \cellcolor{gray!25}{(22)}
    &   \cellcolor{green!25}{1} &     \cellcolor{green!25}{1} &     \cellcolor{green!25}{1}
    &     \cellcolor{green!25}{1} &     \cellcolor{green!25}{1} &     \cellcolor{green!25}{1}
    &     \cellcolor{green!25}{1} &     \cellcolor{green!25}{1} &     \cellcolor{green!25}{1}
    &     \cellcolor{green!25}{1} &     \cellcolor{green!25}{1} &     \cellcolor{green!25}{1}
     &     \cellcolor{green!25}{1} &     \cellcolor{green!25}{1} &     \cellcolor{green!25}{1}
     &     \cellcolor{green!25}{1} &     \cellcolor{green!25}{1} &     \cellcolor{green!25}{1}
      &     \cellcolor{green!25}{1} &     \cellcolor{green!25}{1} &     \cellcolor{yellow!25}{0}
      &     \cellcolor{yellow!25}{0} &    \cellcolor{gray!25}{20}\\
    \Xhline{1.2pt}
\end{tabular}
\end{table*}

\subsection{SFU HDR Database}

The SFU HDR database, recently collected by Funt and
Shi, includes high dynamic range (HDR)
linear images of 105 scenes captured using a Nikon D700 digital still camera.
Three subsets are provided for color constancy research.
The first one includes 105 16-bit base images, with a color
checker placed in the scenes to record the illuminant color.
The second one contains images of the same scene, but with the color checker
removed. The third subset contains 105 images of float format,
which is constructed from the base images.
In the experiment, we use images in the third subset
to evaluate the proposed approach.
For images with a color checker positioned in the scene,
the color checker is masked during illuminant estimation
such that performance of the approach can be fully evaluated.
We track how the the proposed approach performs under varying settings of $\eta$ and plot
the results in Fig. \ref{hdr_vary}.
Again we notice that its performance keeps quite stable while $\eta$ is ranging from 1 to 4.
The results reported in Table \ref{hdr_result} and Table \ref{WST_sfuhdr}
are obtained with $\eta=2$.

Table \ref{hdr_result} shows the results for multiple approaches. In this table, the results of GW, WP,
SoG, GG, GE1 and GE2
were obtained by running the Matlab codes from \cite{cc_website,gijsenij2011computational}
with optimal parameter settings.
The optimal results of DBPCA is reported with $n=1$.
The results of WGE is obtained with the parameter settings $(\kappa,p,\sigma)=(50,1,2)$. 
And the results of WP(post blurred) and
CM are directly cited from \cite{finlayson2013corrected}.
It is noticed that the median error of DCs is lower than most approaches except GG and CM.
Evidently, the performance of DCs on the SFU HDR database
is poorer than that on the SFU laboratory and the Gehler-Shi databases.
Funt and
Shi pointed out in their work \cite{funt2010rehabilitation} that
all scenes in the SFU HDR database contain some variation in the illumination color because of interreflections.
Therefore, interreflections should be undoubtedly responsible
for this poorer performance of DCs on the SFU HDR database.
Table \ref{WST_sfuhdr} shows the results of the sign test. The analysis of CM is again omitted
for its error distribution is unavailable. We notice that
DCs achieves the highest score among the tested algorithms. In particular,
it performs better than
GW, WGE and DOCC(max), while shows indistinguishable performance with the other methods.
The average runtime for each image in this database is 2.65 seconds,
with the time for loading an HDR image neglected.

\begin{table*}[!t]
  \centering
\caption{Performance on the HDR database.}\label{hdr_result}
 \begin{tabular}{|l|c|c|c|C{9mm}|C{9mm}|}
   \hline
   Method & Median & Mean & Trimean & Best-25\%  & Worst-25\% \\
\hline
\hline
   DN & 14.67$^\circ$ & 15.11$^\circ$ & 14.95$^\circ$ & 11.19$^\circ$ & 19.51$^\circ$ \\
\hline
   GW & 7.45$^\circ$ & 8.08$^\circ$ & 7.78$^\circ$ & 1.94$^\circ$ & 15.17$^\circ$ \\
   WGE & 5.76$^\circ$ & 6.67$^\circ$ & 5.72$^\circ$ & 1.78$^\circ$ & 13.29$^\circ$ \\
   WP & 4.06$^\circ$ & 6.36$^\circ$ & 4.75$^\circ$ & 1.59$^\circ$ & 14.45$^\circ$ \\
   WP(blur) & 3.90$^\circ$ & 6.30$^\circ$ & \textbf{--}& \textbf{--} & \textbf{--} \\
   SoG & 3.44$^\circ$ & 5.59$^\circ$ & 4.11$^\circ$ & 1.45$^\circ$ & 12.82$^\circ$ \\
   GG & 3.16$^\circ$ & 5.51$^\circ$ & 3.96$^\circ$ & 1.52$^\circ$ & 12.66$^\circ$ \\
   GE1 & 3.58$^\circ$ & 5.54$^\circ$ & 4.15$^\circ$ & 1.29$^\circ$ & 12.40$^\circ$ \\
   GE2 & 3.71$^\circ$ & 5.60$^\circ$ & 4.42$^\circ$ & 1.31$^\circ$ & 12.33$^\circ$ \\
   DOCC(sum) & 5.27$^\circ$ & 6.43$^\circ$ & 5.60$^\circ$ & 1.59$^\circ$ &13.19$^\circ$ \\
   DOCC(max) & 3.52$^\circ$ & 6.20$^\circ$ & 4.44$^\circ$ & 1.62$^\circ$ & 14.16$^\circ$ \\
   NUS & 3.55$^\circ$ & 5.75$^\circ$ & 4.12$^\circ$ & 1.25$^\circ$ & 13.63$^\circ$\\
   CM(3 edge) & 3.20$^\circ$ & 4.00$^\circ$ & \textbf{--} &\textbf{--} & \textbf{--}\\
   CM(9 edge) & 2.70$^\circ$ & 3.50$^\circ$ & \textbf{--} &\textbf{--} & \textbf{--}\\
\hline
   DCs & 3.37$^\circ$ & 5.56$^\circ$ & 4.11$^\circ$ & 1.27$^\circ$ & 12.86$^\circ$ \\
   \hline
 \end{tabular}
\end{table*}

\begin{table*}[!t]
  \centering
  \small
  \caption{WST test on the SFU HDR database.}\label{WST_sfuhdr}
  \begin{tabular}{!{\vrule width1.2pt}l!{\vrule width1.2pt}c!{\vrule width1.2pt}c
  !{\vrule width1.2pt}c!{\vrule width1.2pt}c!{\vrule width1.2pt}c
  !{\vrule width1.2pt}c!{\vrule width1.2pt}c!{\vrule width1.2pt}c
  !{\vrule width1.2pt}c!{\vrule width1.2pt}c!{\vrule width1.2pt}c
  !{\vrule width1.2pt}c!{\vrule width1.2pt}c!{\vrule width1.2pt}c!{\vrule width1.2pt}}

  \Xhline{1.2pt}
   & \rotatebox{90}{(1)DN}  & \rotatebox{90}{(2)GW} & \rotatebox{90}{(3)WGE}
   & \rotatebox{90}{(4)DOCC(sum)} & \rotatebox{90}{(5)DOCC(max)} & \rotatebox{90}{(6)WP} & \rotatebox{90}{(7)SoG}& \rotatebox{90}{(8)GG} & \rotatebox{90}{(9)GE1}
   & \rotatebox{90}{(10)GE2} & \rotatebox{90}{(11)NUS} & \rotatebox{90}{(12)DCs} & \rotatebox{90}{Score}\\
   \Xhline{1.2pt}
  \cellcolor{gray!25}{(1)}
  &\cellcolor{yellow!25}{0}    &    \cellcolor{red!25}{-1}    &    \cellcolor{red!25}{-1}
  &    \cellcolor{red!25}{-1}    &    \cellcolor{red!25}{-1}    &    \cellcolor{red!25}{-1}
  &    \cellcolor{red!25}{-1}    &    \cellcolor{red!25}{-1}    &    \cellcolor{red!25}{-1}
  &    \cellcolor{red!25}{-1}    &    \cellcolor{red!25}{-1}    &    \cellcolor{red!25}{-1}
  & \cellcolor{gray!25}{0}\\
  \Xhline{1.2pt}
 \cellcolor{gray!25}{(2)}
 &\cellcolor{green!25}{1}     &     \cellcolor{yellow!25}{0}    &     \cellcolor{yellow!25}{0}
 &    \cellcolor{red!25}{-1}    &    \cellcolor{red!25}{-1}    &    \cellcolor{red!25}{-1}
 &    \cellcolor{red!25}{-1}    &    \cellcolor{red!25}{-1}    &    \cellcolor{red!25}{-1}
 &    \cellcolor{red!25}{-1}    &    \cellcolor{red!25}{-1}    &    \cellcolor{red!25}{-1}
 & \cellcolor{gray!25}{1}\\
  \Xhline{1.2pt}
\cellcolor{gray!25}{(3)}
&\cellcolor{green!25}{1}     &     \cellcolor{yellow!25}{0}    &     \cellcolor{yellow!25}{0}
&     \cellcolor{yellow!25}{0}    &     \cellcolor{yellow!25}{0}    &     \cellcolor{yellow!25}{0}
&    \cellcolor{red!25}{-1}    &    \cellcolor{red!25}{-1}    &    \cellcolor{red!25}{-1}
&    \cellcolor{red!25}{-1}    &    \cellcolor{red!25}{-1}    &    \cellcolor{red!25}{-1}
& \cellcolor{gray!25}{1}\\
  \Xhline{1.2pt}
\cellcolor{gray!25}{(4)}
&\cellcolor{green!25}{1}     &     \cellcolor{green!25}{1}     &     \cellcolor{yellow!25}{0}
&     \cellcolor{yellow!25}{0}    &     \cellcolor{yellow!25}{0}    &     \cellcolor{yellow!25}{0}
&     \cellcolor{yellow!25}{0}    &     \cellcolor{yellow!25}{0}    &     \cellcolor{yellow!25}{0}
&     \cellcolor{yellow!25}{0}    &     \cellcolor{yellow!25}{0}    &    \cellcolor{red!25}{-1}
& \cellcolor{gray!25}{2}\\
  \Xhline{1.2pt}
\cellcolor{gray!25}{(5)}
&\cellcolor{green!25}{1}     &     \cellcolor{green!25}{1}     &     \cellcolor{yellow!25}{0}
&     \cellcolor{yellow!25}{0}    &     \cellcolor{yellow!25}{0}    &     \cellcolor{yellow!25}{0}
&     \cellcolor{yellow!25}{0}    &     \cellcolor{yellow!25}{0}    &     \cellcolor{yellow!25}{0}
&     \cellcolor{yellow!25}{0}    &     \cellcolor{yellow!25}{0}    &     \cellcolor{yellow!25}{0}
& \cellcolor{gray!25}{2}\\
  \Xhline{1.2pt}
\cellcolor{gray!25}{(6)}
&\cellcolor{green!25}{1}     &     \cellcolor{green!25}{1}     &     \cellcolor{yellow!25}{0}
&     \cellcolor{yellow!25}{0}    &     \cellcolor{yellow!25}{0}    &     \cellcolor{yellow!25}{0}
&     \cellcolor{yellow!25}{0}    &     \cellcolor{yellow!25}{0}    &     \cellcolor{yellow!25}{0}
&     \cellcolor{yellow!25}{0}    &     \cellcolor{yellow!25}{0}    &     \cellcolor{yellow!25}{0}
& \cellcolor{gray!25}{2}\\
  \Xhline{1.2pt}
 \cellcolor{gray!25}{(7)}
 &\cellcolor{green!25}{1}     &     \cellcolor{green!25}{1}     &     \cellcolor{green!25}{1}
 &     \cellcolor{yellow!25}{0}    &     \cellcolor{yellow!25}{0}    &     \cellcolor{yellow!25}{0}
 &     \cellcolor{yellow!25}{0}    &     \cellcolor{yellow!25}{0}    &     \cellcolor{yellow!25}{0}
 &     \cellcolor{yellow!25}{0}    &     \cellcolor{yellow!25}{0}    &     \cellcolor{yellow!25}{0}
 & \cellcolor{gray!25}{3}\\
  \Xhline{1.2pt}
\cellcolor{gray!25}{(8)}
&\cellcolor{green!25}{1}     &     \cellcolor{green!25}{1}     &     \cellcolor{green!25}{1}
&     \cellcolor{yellow!25}{0}    &     \cellcolor{yellow!25}{0}    &     \cellcolor{yellow!25}{0}
&     \cellcolor{yellow!25}{0}    &     \cellcolor{yellow!25}{0}    &     \cellcolor{yellow!25}{0}
&     \cellcolor{yellow!25}{0}    &     \cellcolor{yellow!25}{0}    &     \cellcolor{yellow!25}{0}
& \cellcolor{gray!25}{3}\\
  \Xhline{1.2pt}
\cellcolor{gray!25}{(9)}
&\cellcolor{green!25}{1}     &     \cellcolor{green!25}{1}     &     \cellcolor{green!25}{1}
&     \cellcolor{yellow!25}{0}    &     \cellcolor{yellow!25}{0}    &     \cellcolor{yellow!25}{0}
&     \cellcolor{yellow!25}{0}    &     \cellcolor{yellow!25}{0}    &     \cellcolor{yellow!25}{0}
&     \cellcolor{yellow!25}{0}    &     \cellcolor{yellow!25}{0}    &     \cellcolor{yellow!25}{0}
& \cellcolor{gray!25}{3}\\
  \Xhline{1.2pt}
\cellcolor{gray!25}{(10)}
&\cellcolor{green!25}{1}     &     \cellcolor{green!25}{1}     &     \cellcolor{green!25}{1}
&     \cellcolor{yellow!25}{0}    &     \cellcolor{yellow!25}{0}    &     \cellcolor{yellow!25}{0}
&     \cellcolor{yellow!25}{0}    &     \cellcolor{yellow!25}{0}    &     \cellcolor{yellow!25}{0}
&     \cellcolor{yellow!25}{0}    &     \cellcolor{yellow!25}{0}    &     \cellcolor{yellow!25}{0}
& \cellcolor{gray!25}{3}\\
  \Xhline{1.2pt}
  \cellcolor{gray!25}{(11)}
&\cellcolor{green!25}{1}     &     \cellcolor{green!25}{1}     &     \cellcolor{green!25}{1}
&     \cellcolor{yellow!25}{0}    &     \cellcolor{yellow!25}{0}    &     \cellcolor{yellow!25}{0}
&     \cellcolor{yellow!25}{0}    &     \cellcolor{yellow!25}{0}    &     \cellcolor{yellow!25}{0}
&     \cellcolor{yellow!25}{0}    &     \cellcolor{yellow!25}{0}    &     \cellcolor{yellow!25}{0}
& \cellcolor{gray!25}{3}\\
  \Xhline{1.2pt}
 \cellcolor{gray!25}{(12)}
&\cellcolor{green!25}{1}     &     \cellcolor{green!25}{1}     &     \cellcolor{green!25}{1}
&     \cellcolor{green!25}{1}     &     \cellcolor{yellow!25}{0}    &     \cellcolor{yellow!25}{0}
&     \cellcolor{yellow!25}{0}    &     \cellcolor{yellow!25}{0}    &     \cellcolor{yellow!25}{0}
&     \cellcolor{yellow!25}{0}    &     \cellcolor{yellow!25}{0}    &     \cellcolor{yellow!25}{0}
& \cellcolor{gray!25}{4}\\
  \Xhline{1.2pt}
\end{tabular}
\end{table*}

\begin{table*}[!t]
  \centering
\caption{Inter-database performance test.}\label{inter}
 \begin{tabular}{!{\vrule width1.2pt}c!{\vrule width1.2pt}c!{\vrule width1.2pt}c
 !{\vrule width1.2pt}c!{\vrule width1.2pt}c!{\vrule width1.2pt}c
 !{\vrule width1.2pt}c!{\vrule width1.2pt}c!{\vrule width1.2pt}}
   \Xhline{1.2pt}
  \multicolumn{2}{!{\vrule width1.2pt}c!{\vrule width1.2pt}}{\multirow{2}{*}{Method}} & \multicolumn{2}{c!{\vrule width1.2pt}}{SFU laboratory} &  \multicolumn{2}{c!{\vrule width1.2pt}}{Gehler-Shi} & \multicolumn{2}{c!{\vrule width1.2pt}}{SFU HDR} \\
   \Xcline{3-8}{1.2pt}
   \multicolumn{2}{!{\vrule width1.2pt}c!{\vrule width1.2pt}}{} & Median & Mean & Median & Mean & Median & Mean \\
\Xhline{1.2pt}
\multirow{3}{*}{DCs} & $\eta=2$ & 1.71$^\circ$ & 4.21$^\circ$ & 2.09$^\circ$ & 3.52$^\circ$ & 3.37$^\circ$ & 5.56$^\circ$ \\
\Xcline{2-8}{1.2pt}
& $\eta=3$ &  1.82$^\circ$ & 4.15$^\circ$ & 1.98$^\circ$ & 3.24$^\circ$ & 3.35$^\circ$ & 5.76$^\circ$ \\
\Xcline{2-8}{1.2pt}
& $\eta=4$ & 2.11$^\circ$ & 4.35$^\circ$ & 1.86$^\circ$ & 3.14$^\circ$ & 3.20$^\circ$ & 5.72$^\circ$   \\
\Xhline{1.2pt}
\multicolumn{2}{!{\vrule width1.2pt}c!{\vrule width1.2pt}}{DOCC(max)} & 2.61$^\circ$ & 5.41$^\circ$ & 3.30$^\circ$ & 5.76$^\circ$ & 3.55$^\circ$ & 5.80$^\circ$ \\
\Xhline{1.2pt}
 \end{tabular}
\end{table*}

\subsection{Inter-database Evaluation}\label{inter-database}
Inter-database performance of an approach is very important to ensure
its general effectiveness on different images. Recently,
this important characteristics is explored by two approaches, Exemplar \cite{vaezi2014exemplar} and DOCC \cite{gao2015color}.
Particularly, Exemplar provides its inter-database results by training the model on one database and testing it on another one,
while the results of DOCC are obtained by keeping the parameters to be constant across different databases. Similar to DOCC,
 we report the inter-database results of DCs by setting $\eta$ to be identical on different databases.
 Although the inter-database performance of DOCC is more competitive compared to that of Exemplar,
the stable performance of DCs makes it achieve
far superior inter-database results to DOCC.
We show such test results of DCs in Table \ref{inter}, and specifically for $\eta=2, 3, 4$.
Besides, we add the best performance of DOCC under inter-database test
in the last row for better comparison. It is noticed that the inter-database performance of DCs
is generally much better than that of DOCC. As a matter of fact, the results of DCs under inter-database configurations
are even superior to the optimal results of DOCC on each database. Yet, the most exciting
robustness of DCs should be its surprisingly stable performance in WST test. No methods outperform DCs on each database whether the parameter $\eta$ is
2, 3, or 4. In details, on the SFU laboratory database, its WST test remains the same for $\eta=3$. And for $\eta=4$, DCs just no longer performs better than WGE.
 On the Gehler-Shi database, its WST test remains the same for both $\eta=2$ and $\eta=3$. And on the SFU HDR database, DCs just no longer outperforms
 DOCC(max) for either $\eta=3$ or $\eta=4$.
In practice,
$\eta=2$ is suggested since the time cost for processing an image will be decreased,
and meanwhile the performance is quite satisfactory.

\section{Discussion}
We discuss connections between DCs and works based on gray-edge in this section. According to our analysis,
the gray-edge assumption is possible to hold only in achromatic regions and uniform highlight regions.
Therefore, the improvement achieved by \cite{joze2012role} and WGE \cite{gijsenij2012improving} should be contributed to the substantial elimination of
negative influence caused by derivatives of pixels beyond these useful regions. In \cite{gijsenij2012improving}, Gijsenij et al.
evaluated the original gray-edge method \cite{van2007using} on three kinds of edge types: specular, shadow and material.
One thing which should be noted explicitly is that the edge classification mechanism they adopt makes
derivatives of achromatic pixels to be both specular edges and shadow edges. As a result, we speculate that in their evaluations, the acceptable performance of shadow edges compared to
material edges is due to the high weights assigned to derivatives of pixels in achromatic regions.
And the better performance of specular edges is attributed to the high weights assigned to derivatives of pixels in both achromatic and uniform highlight regions.
Nevertheless, all of these gray-edge methods (i.e., \cite{gijsenij2012improving, joze2012role, van2007using}) shows poor inter-database performance, which consequently makes them inferior to the fine and stable performance of DCs. As to DBPCA \cite{cheng2014illuminant}, its effectiveness should be dominantly contributed to the achromatic pixels. However, the main drawback of DBPCA is that it cannot exploit the useful specular component in highlight regions.

\section{Conclusion}
 In this paper, we mainly show that under the NIR assumption on the dichromatic reflection model,
 derivative colors with large ratio $\frac{|\triangle m_s|}{|\triangle m_d|}$ in highlight regions approximate the
 illuminant color quite well. Besides, achromatic regions can provide derivative colors with the $\frac{|\triangle m_s|}{|\triangle m_d|}$ to be $\infty$, under the neutral illuminant assumption.
 And these derivative colors are satisfactorily extracted from achromatic regions and uniform regions with strong specularity
 in real images, using the second-order gaussian functions as the differential operators.
 Although it is impossible for all extracted derivative colors to have large ratio $\frac{|\triangle m_s|}{|\triangle m_d|}$,
 the majority of them are densely distributed around the ground truth in the rg chromaticity space.
 Therefore, kernel density estimation is chosen to estimate the illuminant color from the extracted derivative colors.
 Despite the simplicity of the proposed approach,
our experiments on three standard databases demonstrate its feasibility and fine performance,
 which is either better or quite competitive in comparison to the existing algorithms. More significantly, the inter-database performance of our approach
 is quite pleasant and superior to those of state-of-the-art methods.
The fine performance of the proposed approach demonstrates that the physics-based method, commonly accused of mediocre performance in the past, is able to perform either better than or comparable to the complex statistics-based methods.
 %

\section*{Acknowledgement}
This work was partially supported by the Research Grants Council General Research Fund under Grant 618711, the National Basic Research Program of China under Grant 2012CB316300, the 111 Project of China under Grant B08038, and the Natural Science Foundation of China under Grant 61403292.


\end{document}